\documentclass[Afour,sageh,times]{sagej}

\usepackage{moreverb,url}
\usepackage{tikz}
\usetikzlibrary{bayesnet}
\usepackage{subfigure}
\usepackage{multicol, blindtext}
\usepackage{nameref}

\usepackage[colorlinks,bookmarksopen,bookmarksnumbered,citecolor=red,urlcolor=red]{hyperref}

\newcommand\BibTeX{{\rmfamily B\kern-.05em \textsc{i\kern-.025em b}\kern-.08em
T\kern-.1667em\lower.7ex\hbox{E}\kern-.125emX}}

\def\volumeyear{2016}

\begin{document}

\runninghead{Learning Generative Visual Control}

\title{Learning a generative model for robot control using visual feedback}

\author{Nishad Gothoskar\affilnum{1}, Miguel L\'azaro-Gredilla\affilnum{1}, Abhishek Agarwal\affilnum{1}, Yasemin Bekiroglu\affilnum{1}, Dileep George\affilnum{1}}

\affiliation{\affilnum{1}Vicarious AI}

\corrauth{Dileep George, Vicarious AI, Union City, CA 94587}

\email{dileep@vicarious.com}

\begin{abstract}
We introduce a novel formulation for incorporating visual feedback in controlling robots. We define a generative model from actions to image observations of features on the end-effector. Inference in the model allows us to infer the robot state corresponding to target locations of the features. This, in turn, guides motion of the robot and allows for matching the target locations of the features in significantly fewer steps than state-of-the-art visual servoing methods. The training procedure for our model enables effective learning of the kinematics, feature structure, and camera parameters, simultaneously. This can be done with no prior information about the robot, structure, and cameras that observe it. Learning is done sample-efficiently and shows strong generalization to test data. Since our formulation is modular, we can modify components of our setup, like cameras and objects, and relearn them quickly online.  Our method can handle noise in the observed state and noise in the controllers that we interact with. We demonstrate the effectiveness of our method by executing grasping and tight-fit insertions on robots with inaccurate controllers.
\end{abstract}


\maketitle

\begin{abstract}

\end{abstract}
\section{Introduction}
\label{sec:introduction}

A main objective of developing robots is to enable them to plan and execute tasks. Often, the tasks we want robots to perform involve interacting with objects through grasping, inserting, etc. An intelligent robotic system should be able to flexibly operate in noisy and uncertain real-world environments by integrating information from various sensory modalities. The human sensorimotor loop receives input from a variety of sensory sources and produces motor commands. An important sensory source in this loop is visual perception, which provides rich information about the surrounding environment. Visual input is important for humans to interact with the world around them and there is substantial experimental evidence of visuomotor coupling across many regions of the brain \citep{attinger2017visuomotor}. This coupling allows us to foveate to and track visual features and then manipulate objects using our limbs and tools.

In robotics, techniques using visual feedback to control robots are referred to as visual servoing. There are two main methods of visual servoing: Image-based visual servoing (IBVS) and Position-based visual servoing (PBVS). IBVS extracts visual features, and then the control law minimizes the errors between the current and desired feature locations in the image. PBVS uses the extracted visual features to infer the pose of the observed object, and then the control law minimizes the error between the current and desired pose. In general, the goal of these methods is to achieve a desired position of the tracked object using information extracted from the image.

Standard methods for visual servoing require calibration of the robot kinematics and the cameras. Kinematic calibration often requires a precisely manufactured device or a laser-based depth sensor for providing accurate measurements of the pose of the end-effector. Camera calibration often requires a routine using a checkerboard placed at different locations in view of the camera to calibrate the intrinsics and extrinsics parameters. These procedures must be repeated each time a change is made to the robot's structure or the camera positions.

This is in stark contrast to the flexibility with which humans can perform control. Consider if instead of the visual servoing algorithm, we allow a human to operate a joystick controlling the robot's joints. By observing how the robot moves in response to the joystick, the human would quickly learn to control the robot. They would be robust to changes in perspective or modifications to the structure of the robot. To replicate this ability in an algorithm, we need to consider the perception and control problems simultaneously, so that we can adapt our visual model or control policy in response to changes.

In this paper, we present a method for learning to control a robot from visual feedback. We learn a generative model from actions (sent to a joint controller or Cartesian controller) to feature pixel coordinates in an image. The model can be learned with no prior information about the robot, features, or camera. Inference in this model allows us to accurately position the robot end-effector using those visual features. This circumvents the need for explicit kinematic and camera calibration procedures. Additionally, it allows us to be unaffected by the errors that may be introduced by inaccurate calibrations. We can learn the parameters of this model with few samples and can continue to learn online, allowing our system to be deployed quickly and be robust to changes to the setup.

\subsection{Related Work}

\subsubsection{Image-based visual servoing} Most methods for image-based visual servoing attempt to estimate the image Jacobian. The image Jacobian governs how changes in the robot state (joint state or Cartesian pose) will produce changes to the features' pixel coordinates in the image. Then, given this Jacobian and the current and desired feature pixel coordinates, we can backpropogate the errors in the pixel space to errors in the robot's state space. This error in the state space is used to derive a command to modify the robot's state. \cite{kragic2002survey} provides a more complete survey of the variety of methods for visual servoing, in the context of manipulation tasks.

The commonly used software package VISP \citep{marchand2005visp} implements various methods for visual servoing, including for an eye-to-hand setup. The control law they use is as follows:

\begin{equation}
\label{eq:visp}
\begin{aligned}
\mathbf{\dot{q}} = \gamma \mathbf{\Big( \hat{L}_e {}^{c} V_e {}^{e} J_e(q) \Big)^+ e}
\end{aligned}
\end{equation}
$\mathbf{{}^{e} J_e (q)}$ is the robot Jacobian, expressed in the end-effector frame, which is a function of the joint states $q$. $\mathbf{{}^{c} V_e}$ defines the transform between the end-effector frame and the camera frame. $\mathbf{\hat{L}_e}$ is the interaction matrix which captures how changes to the 3D positions of the features relate to the changes in their pixel locations. The error between the current and desired coordinates is $e$, where coordinates include both pixel coordinates $x$ and $y$ and the distance from the camera origin $z$. With this control law, we compute $\dot{q}$,  the velocity command to send to the joints. $\gamma$ is a gain term that controls the step size of this command and allows for exponential convergence of the error. 

To find $\mathbf{{}^{e} J_e (q)}$, we must have access to the robot kinematic model and the joint states. $\mathbf{{}^{c} V_e}$ is the output of a camera extrinsics calibration. The interaction matrix $\mathbf{\hat{L}_e}$ is approximated online. Our model is similar to VISP in terms of its structure, however \emph{while VISP requires the above values to be provided, our method is able to learn them from data}.

\subsubsection{Independent camera and kinematics calibration}
Previous work has proposed methods to calibrate the kinematics and camera parameters. \cite{tsai1989new} present a method for calibrating a eye-to-hand setup, along with methods for camera calibration \citep{tsai1980review} and kinematic calibration \citep{lenz1989calibrating}. While this three-stage procedure computes the same parameters as our model, they require a specific setup to do so. They require a calibration board and accurate readings of the joint states. In addition, they must individually move each of the joints during the kinematic calibration procedure. \emph{Our method does not have either of these constraints and is robust to noisy joint state measurements.}

Camera calibration is a deeply studied topic. \cite{zhang2000flexible} proposed a technique for calibration using a planar pattern observed from multiple viewpoints. This algorithm has been implemented in the software packages Robot Operating System (ROS) \citep{quigley2009ros} and OpenCV \citep{bradski2000opencv} and is a commonly used method for calibration in robotic systems. This method also requires a calibration board. And further, to use this method for an external camera (a camera not attached to the robot arm), an on arm camera is also needed to simultaneously view the calibration board. In our method, we use many of the same techniques as are used in camera calibration. However we are able to do so \emph{with only the external cameras and without the need for a calibration board}.

For camera intrinsics calibration, in a step away from requiring a calibration board, \cite{fraser1997digital} and \cite{faugeras1992camera} propose methods for self-calibration that are able to identify the intrinsics parameters by tracking features as they move across an image. \cite{hemayed2003survey} provides a survey of other useful methods for self-calibration. In this work, we do not consider the identification of intrinsics parameters, since the factory-provided calibration served as a sufficiently good initialization for our optimization procedure. If this were not the case, we would also apply these self-calibration methods.

In kinematic calibration, there has been work on attempting to learn a robot's kinematics model. \cite{hollerbach1996calibration} provides a survey of the methods for open-loop and closed-loop kinematic calibration. \cite{ikits1997kinematic} uses a plane constraint on the robot endpoint to make it feasible to solve for the kinematic parameters. \cite{renders1991kinematic} proposes a calibration method based on a maximum likelihood approach to identifying model errors, using a device for accurately measuring the length of motions by the end-effector.

Motivated by the need for a computationally efficient solution for inverse kinematics, \cite{d2001learning} developed a spatially localized learning approach for inverting the kinematics. Connecting kinematics learning to human motor learning,  \cite{rolf2010goal} shows how goal babbling can be used to learn an inverse kinematics model from few samples.  \cite{atkeson1989learning} also attempts to understand the information processing issues involved in motor learning and evaluating their biological plausibility. In all of the above work on kinematic learning, either the states of the joints or the pose of the end-effector are observed and used for learning. Our method is able to be learn \emph{with only observations of the commands sent to a noisy controller and images}. In addition, our method \emph{does not require any measurement device or constraint on the data collection}, as it can be learned from random motions.

\subsubsection{Model-free approaches}
Another class of methods for visual servoing are uncalibrated methods. The uncalibrated methods directly estimate the image Jacobian and use it for guiding servoing.

For the setting of a calibrated robot but uncalibrated cameras, a variety of methods have been proposed. \cite{yoshimi1994active} leverages the fact that features will trace ellipses on the image as the camera is rotated. However, this method heavily relies heavily on being able to manipulate an axis of the camera, which is not applicable to our setup with an external camera. \cite{liu2006uncalibrated} proposes a depth-independent interaction matrix which is an improvement on the types of methods implemented in VISP that require both pixel coordinates and depth values. \cite{malis2004visual} proposed a visual servoing method that is invariant to changes in the camera intrinsic parameters. Recent work by \cite{shademan2010robust} has made these procedures more robust to outliers and noise using an M-estimator. 

There has also been previous work on visual servoing when both the robot and cameras are uncalibrated, the setting that we are intersted in. \cite{hosoda1994versatile} and \cite{hosoda1998adaptive} have presented methods for uncalibrated visual control for static targets and fixed cameras using Broyden updating of the Jacobian estimate.  \cite{piepmeier2004uncalibrated} uses a dynamic quasi-Newton method for Jacobian estimation and extends on the previously stated works by tracking a moving target.

Our work departs from these model-free, uncalibrated methods. These model-free methods learn local differential models while our method learns a global model. Model-free methods will always have an increased sample complexity, which will make learning slower and will only be beneficial in setups outside our scope (i.e., control of elements that are not robotic arms).

\subsubsection{Joint camera and kinematics calibration}
The previous work most related to our proposed method are those which perform a global optimization over the kinematic and camera parameters. \cite{puskorius1987global} does this optimization, but requires tracking of a fixed calibration point from multiple viewpoints. \cite{zhuang1995simultaneous} improves on this procedure but still requires a good initialization of the parameters to be provided. Both these methods need to observe the joint states of the robot. Our method is able to perform the global optimization with no prior information and can be done without directly observing the joint states, only the actions sent to a noisy controller.

\subsubsection{Deep reinforcement learning of control}
Recent work has explored learning control using deep neural networks. Deep reinforcement learning has been successfully applied to a variety of domains, as reviewed in \citep{arulkumaran2017deep}, including continuous-action domains \citep{lillicrap2015continuous}. The authors of \citep{levine2016end} study learning of perception and control, end-to-end. They design a policy search algorithm and a convolutional neural network (CNN) architecture to learn a policy that produces joint torques given image observations. A similar method is used in \citep{levine2018learning}, to learn hand-eye coordination to perform grasping tasks. A model-based approach to this problem is taken in \citep{finn2017deep} by pairing a CNN, which predicts future image frames given the current image and control signals, with a model-predictive controller. In general compared to these neural-network based methods, our model has significantly fewer parameters and therefore requires much less training data to learn. Our method, like \cite{finn2017deep}, learns a generative model, however does so with a specialized modular formulation rather than using general function approximation. This endows our approach with a much improved sample efficiency, strong generalization to test data, and the ability to quickly adapt to new structural setups, for example when cameras and objects are changed.

\subsection{Contributions}

The contributions of our work are:
\begin{itemize}
\item We propose a learning procedure that enables consistent convergence of end-to-end learning of an arbitrary, uncalibrated robot observed by an arbitrary number of uncalibrated cameras, for which no other specific information is available.
\item Our global formulation enables us to match target locations of image features in significantly fewer iterations than state-of-the-art methods in visual servoing.
\item By explicitly modeling noise, we can accurately control noisy robots. We demonstrate this by performing grasping and tight-fit insertion on a Baxter robot with an inaccurate controller.
\item To the best of our knowledge, ours is the first method that can learn accurate control from only observing the actions sent to a noisy controller and image features. Previous methods require observations of the robot's joint states.
\item We present quantitative measures of the relative difficulties of learning kinematic, structure, and camera parameters.
\end{itemize}
\section{Model}
\label{sec:model}

\begin{figure*}
\setlength{\fboxsep}{0pt}
\setlength{\fboxrule}{0pt}
  \centering
      \tikz[scale=0.6, every node/.style={scale=0.8}]{ %
        \node[latent] (j_t) {$\mathbf{j}_t$} ;
        \node[latent, right=of j_t] (p_t) {$P_t$} ;
  		\edge {j_t} {p_t} ;
        \node[latent, right=of p_t] (m_t) {$M_t$} ;
  		\edge {p_t} {m_t} ; %
        \node[obs, right=of m_t] (i_t) {$I_t$} ;
  		\edge {m_t} {i_t} ;
  
        \node[latent, below=of j_t] (j_tp1) {$\mathbf{j}_{t+1}$} ;
        \node[latent, right=of j_tp1] (p_tp1) {$P_{t+1}$} ;
  		\edge {j_tp1} {p_tp1} ;
        \node[latent, right=of p_tp1] (m_tp1) {$M_{t+1}$} ;
  		\edge {p_tp1} {m_tp1} ; %
        \node[obs, right=of m_tp1] (i_tp1) {$I_{t+1}$} ;
  		\edge {m_tp1} {i_tp1} ;
  		
        \node[obs, left=of j_t] (a_t) {$\mathbf{a}_t$} ;
  		\edge {j_t} {j_tp1} ;
  		\edge {a_t} {j_tp1} ;

        \node[obs, left=of j_tp1] (a_tp1) {$\mathbf{a}_{t+1}$} ;

	    \node[latent, below=of j_tp1] (j_tp2) {$j_{t+2}$} ;
  		\edge {j_tp1} {j_tp2} ;
  		\edge {a_tp1} {j_tp2} ;

      }
  \caption{Bayesian network describing the generative model of the features observed in an image from the actions sent to a joint delta controller. Shaded nodes are observed. All the other nodes are unobserved, except as detailed in some of the experiments.}
  \label{fig:joint_gm}
\setlength{\fboxsep}{0pt}
\setlength{\fboxrule}{0pt}
  \centering
      \tikz[scale=0.6, every node/.style={scale=0.8}]{ %
        \node[latent] (p_t) {$P_t$} ;
        \node[latent, right=of p_t] (m_t) {$M_t$} ;
  		\edge {p_t} {m_t} ; %
        \node[obs, right=of m_t] (i_t) {$I_t$} ;
  		\edge {m_t} {i_t} ;
  
        \node[latent, below=of p_t] (p_tp1) {$P_{t+1}$} ;
        \node[latent, right=of p_tp1] (m_tp1) {$M_{t+1}$} ;
  		\edge {p_tp1} {m_tp1} ; %
        \node[obs, right=of m_tp1] (i_tp1) {$I_{t+1}$} ;
  		\edge {m_tp1} {i_tp1} ;
  		
        \node[obs, left=of p_t] (a_t) {$A_t$} ;
  		\edge {p_t} {p_tp1} ;
  		\edge {a_t} {p_tp1} ;

        \node[obs, left=of p_tp1] (a_tp1) {$A_{t+1}$} ;

	    \node[latent, below=of p_tp1] (j_tp2) {$p_{t+2}$} ;
  		\edge {j_tp1} {j_tp2} ;
  		\edge {a_tp1} {j_tp2} ;

      }
  \caption{Bayesian network describing the generative model of the features observed in an image from the actions sent to a Cartesian controller. Shaded nodes are observed. All the other nodes are unobserved, except as detailed in some of the experiments.}
  \label{fig:cart_gm}
\end{figure*}

Our method centers around learning the generative processes depicted in Figure \ref{fig:joint_gm} and Figure \ref{fig:cart_gm}. The nodes of the model correspond to the following values:
\begin{itemize}
\item$\mathbf{j}_t$ - joint states at time $t$
\item$P_t$ - Cartesian pose of the end-effector in the origin frame at time $t$
\item$M_t$ - 3D coordinates of the feature points in the origin frame at time $t$
\item$I_t$ - features' pixel coordinates in the images of each camera at time $t$
\end{itemize}

In Figure \ref{fig:joint_gm}, $\mathbf{a}_t$ corresponds to the desired delta of the joint states that is sent to a joint controller. In Figure \ref{fig:cart_gm}, $A_t$ corresponds to a desired transformation of the end-effector pose that is sent to a Cartesian controller. We assume the image is always observed, but in the variety of settings we consider in the following sections, a different combination of the remaining nodes will be observed and unobserved. The final setting we consider is one where only the actions and image are observed, as shown in these figures.

In the following sections, we describe the components of the model and their specific parametrizations. 

\subsection{Kinematics}

Forward kinematics govern how joint angles produce a Cartesian pose of the end-effector. The forward kinematics are robot specific as they depend on the robot's link lengths, rotation of the joints, and relative transformations between joints. For our model, we parametrize the robot kinematics using the Denavit–Hartenberg (DH) convention \citep{hartenberg1964kinematic}. For each joint, there are 4 DH parameters: 

\begin{itemize}
\item$\omega$ - joint angle offset
\item$d$ - link offset
\item$a$ -  link length
\item$\alpha$ - link twist
\end{itemize}

These 4 parameters along with the state of the joint define a homogeneous transformation corresponding to that joint:

\begin{gather*}
T_{\omega, d, a, \alpha} (j) = \\ \scalebox{0.75}{$\left[
\begin{array}{ccc|c}
    \cos(\omega + j) & -\sin(\omega + j) \cos\alpha & \sin(\omega + j) \sin\alpha & a \cos(\omega + j) \\
    \sin(\omega + j) & \cos(\omega + j) \cos\alpha & -\cos(\omega + j) \sin\alpha & a \sin(\omega + j) \\
    0 & \sin\alpha & \cos\alpha & d \\
    \hline
    0 & 0 & 0 & 1
  \end{array}
\right] $}
\end{gather*}

When we compose these transformation for all joints of the robot, we get a transformation to the end-effector frame from the base frame. The base frame of the robot does not necessarily need to correspond to the origin frame. We additionally parametrize the 6D pose of the base frame of the robot as the transformation $B$. 

We collect the DH parameters for all joints into the vectors $\boldsymbol\omega, \mathbf{d}, \mathbf{a}, \boldsymbol\alpha$. The state of each joint at time $t$ is contained in the vector $\mathbf{j}_t$ which is of length $n$ (degrees of freedom of the robot). $\mathbf{j}_t[i]$ is the state of joint $i$. The forward kinematics, which represents the transformation to the end-effector frame from the origin frame, is defined as follows:

\[
P_t = K_{B,\boldsymbol\omega, \mathbf{d}, \mathbf{a}, \boldsymbol\alpha}(\mathbf{j}_t) = B \prod_{i = 1}^{n} T_{\boldsymbol\omega_i, \mathbf{d}_i, \mathbf{a}_i, \boldsymbol\alpha_i} (\mathbf{j}_t[i])
\]

We choose the DH parametrization specifically because it is complete to describe arbitrary robot kinematics. It is also a minimal representation and only requires simple constraints, like positivity, that are easy to enforce. A disadvantage to note is that DH parameters are singular when adjacent joint axes are parallel, since there will be large changes to $d$ for even small deviations in the axis orientations. Most robots do not have parallel axes so this problem does not arise. For such robots we can still apply the presented method but will use a non-minimal representation with constraints.

Henceforth, we will refer to all the kinematics parameters $\boldsymbol\omega, \mathbf{d}, \mathbf{a}, \boldsymbol\alpha$ and base frame parameter $B$ collectively as $R$.

\subsection{Structure}
In order to find the correspondence between the robot's configurations and the observed images, we track features that are rigidly attached to the end-effector. Since features are points, we can parametrize them by their 3D coordinates. Note that these are relative coordinates defined in the end-effector frame. We collect these 3D coordinates coordinates for $m$ feature points into the matrix $F$. Then, given the homogeneous matrix $P$, corresponding to the pose of the end effector in origin frame, the coordinates of the features points in the origin frame are computed as follows:

\[
M_t = S_{F}(P_t) = P_t  \left[ \begin{array}{cccc}
    F_{1x} & F_{1y} & F_{1z} & 1 \\
    F_{2x} & F_{2y} & F_{2z} & 1 \\
     \vdots & \vdots & \vdots & \vdots\\
    F_{mx} & F_{my} & F_{mz} & 1 \\
  \end{array} \right]^T
\]

While here we have assumed the features are attached to the end-effector, these features can be attached to any of the joints of the arm. More coverage of the joints will allow for better inference of the robot's configuration. For the features that are attached to other joints, $P$ will be the Cartesian pose of that joint and $F$ will contain the relative coordinates in that joint's reference frame. Though if we seek to accurately control the end-effector then we must track features on the end-effector.

By learning the feature's coordinates, we are learning the structure of an object attached to the arm. Therefore, many of the same techniques for learning structure from multiple viewpoints can be applied for our model. 

\subsection{Cameras}
For the projection of these features onto the image, we assume a pinhole camera model. This model consists of the extrinsics matrix that is derived from the camera's pose in the world and the intrinsics matrix that defines how 3D points in the camera frame are projected onto the image plane at certain pixels. The extrinsics matrix is a rigid transformation that is the inverse of the camera pose in the origin frame. It is parametrized by $R$ and $t$ which define a 6D pose. The intrinsics matrix is defined by the focal length in $x$ and $y$ ($f_x$ and $f_y$) and the principal point in $x$ and $y$ ($p_x$ and $p_y$). There are additional distortion parameters that can be modeled, but we observed that this had little effect on the performance of our method. The extrinsics matrix $E$ and intrinsics matrix $K$ are defined as follows:

\begin{equation*}
\label{eq1}
\begin{split}
K = \left[
\begin{array}{cccc}
    f_x & 0 & p_x & 0  \\
    0 & f_y & p_y & 0  \\
    0 & 0 & 1 & 0  \\
  \end{array}
\right]
\end{split}
\end{equation*}

\begin{equation*}
\begin{split}
E = \left [ \begin{array}{c|c} \mathbf{R} & \mathbf{t} \\ \hline \mathbf{0} & 1 \end{array} \right ]
\end{split}
\end{equation*}

The 3D coordinates of $m$ features in the origin frame are collected into the matrix $M$. The projection of those points to pixel locations is as follows:

\begin{equation*}
\begin{split}
I_t = C_{K,E}(M_t) = K \ E  \ \left[ \begin{array}{cccc}
    M_{t1x} & M_{t1y} & M_{t1z} & 1 \\
    M_{t2x} & M_{t2y} & M_{t2z} & 1 \\
     \vdots & \vdots & \vdots & \vdots\\
    M_{tmx} & M_{tmy} & M_{tmz} & 1 \\
  \end{array} \right]^T
\end{split}
\end{equation*}

Note that the projection onto the image plane, which transforms from 3D to 2D, requires a normalization of the homogeneous vector and therefore is a nonlinear operation.

Here we have shown the camera projection for a single camera. When there are multiple cameras, we will have different intrinsics and extrinsics matrices for each camera.  We collect the intrinsics and extrinsics parameters for a single camera into $V$. We then have $V_1$, $V_2$...$V_c$ for the $c$ different cameras.

\subsection{Summary}

\begin{equation*}
\begin{split}
P_t &= K_{R}(\mathbf{j}_t) \\
M_t &= S_{F}(P_t) \\
I_t &= C_{V}(M_t) \\
\end{split}
\end{equation*}

For completeness, we summarize how each of the modules described in the above sections map to the generative models in Figures \ref{fig:joint_gm} and \ref{fig:cart_gm}. Given the joint states $\mathbf{j}_t$, the forward kinematics model (defined by $R$) will produce a Cartesian pose of the end-effector $P_t$ in the origin frame. The features are at relative coordinates $F$ in the end-effector frame so when they are applied to the pose $P_t$ it will produce the 3D coordinates $M_t$ of the features in the origin frame. Finally, these coordinates $M_t$ are projected onto camera $i$ (defined by $V_i$) and produce an image where those features are at pixel locations $I_{ti}$.

For an $n$ degree of freedom robot, with $m$ features tracked, observed in $c$ cameras, the number of parameters of our model is: 

\begin{itemize}
  \item Robot base frame: $6$
  \item DH parameters for $n$ joints: $4n$
  \item Structure coordinates for $m$ features: $3m$
  \item Extrinsics for $c$ cameras: $6c$
  \item Intrinsics for $c$ cameras: $4c$
\end{itemize}

Compared to deep neural networks, this model has significantly fewer parameters. As we will discuss in later sections, this is important distinction that gives improved sample-efficiency and generalization.

\section{Learning from Joint States and Images}
\label{sec:learning}

In this section, we summarize the learning procedure when the available data is images and the corresponding joint states. In this setting, we do not need to consider the actions because we assume that if we can read the joint states then we can also command the robot by directly sending joint states.

\subsection{Camera and Structure Parameter Learning} \label{sec:camstructlearning}
Given images in which we have detected the features' pixel locations (in $c$ cameras), we seek to learn the camera parameters for each camera, the features' relative 3D coordinates in the end-effector frame, and the Cartesian pose of the end-effector at each timestep. Note that the Cartesian pose is not a parameter of the model but rather a node in the generative model. We have found that this truncated model, which only considers the poses, 3D points, and projections, is feasible to learn, independent of the joint states and kinematic parameters. It however provides less precise inference of the end-effector 6D pose. We discuss this in further detail in a later section.

The learning procedure simultaneously optimizes the above mentioned parameters. The function $D_{\text{pixel}}(g,a)$ is a quadratic error between the generated and actual pixel coordinates, which is computed only over the features that are observed in the actual image. This error is commonly referred to as the reprojection error. The optimization is:

\begin{equation}
\label{eq:losscamstruct}
\begin{gathered}
L_{\text{cam}}(P_{1:T}, F, V_{1:c}, I_{1:T}) = \\
\sum_{t=1}^{T} \sum_{i=1}^{c} D_{\text{pixel}}(C_{V}(S_F(P_t)), I_{ti})
\end{gathered}
\end{equation}
\begin{equation}
\label{eq:optimcamstruct}
\begin{gathered}
P_{1:T}^*, F^*, V_{1:c}^* = \\ \underset{P_{1:T}, F, V_{1:c}}{\text{argmin}} \ L_{\text{cam}}(P_{1:T}, F, V_{1:c}, I_{1:T}^{o})
\end{gathered}
\end{equation}
We assume we have $c$ cameras and have collected data at $T$ timesteps. $P_{1:T}$ are the transformation matrices of the end-effector's pose at each timestep. $V_{1:c}$ are the extrinsics and intrinsics parameters of each of the $c$ cameras. $I_{ti}$ contains the pixel coordinates of the features in camera $i$ at timestep $t$. The loss in Eq. \eqref{eq:losscamstruct} is the error between pixel coordinates contained in $I_{1:T}$ and the pixel coordinates generated from poses $P_{1:T}$ through our partial model. In, Eq. \eqref{eq:optimcamstruct}, the superscipt $o$ on $I_{1:T}^o$ indicates that these pixel locations are observed and the observed values are used in the optimization.

However, this learning is prone to convergence to sub-optimal local minima and, as such, it is important to provide a good initialization for this optimization. The following sections describe two different methods for computing a good initialization of $P_{1:T}$ and $V_{1:c}$. We have found that a random initialization of $F$ is sufficient and does not affect convergence.

\subsection{Initialization by Triangulation}
If the cameras are placed such that in many timesteps features are observed in multiple cameras, we can leverage this to compute an initialization for our optimization. With 5 feature correspondences between cameras, we can estimate the fundamental matrix between the cameras \citep{nister2004efficient}. This means we only need 5 timesteps at which a feature was observed in both cameras. From the fundamental matrix and an estimate of the intrinsic parameters (we use the factory-provided intrinsic estimate) we can get the essential matrix. Then from this essential matrix, we can recover 4 possible solutions for the camera baseline, the transformation between the two cameras. To select one of these 4 solutions, we triangulate the observed points to 3D and select the configuration of the cameras that places the most observed features in front of both cameras.

This procedure can be repeated for pairs of cameras that observe 5 of the same features until we have accounted for all the cameras. If a camera is unable to be added through, this procedure because it does not observe sufficiently many features, then we can ignore it from the initial optimization and reincorporate it later in the learning procedure.

With this estimate of the poses of the cameras, we can triangulate all the features observed in multiple cameras to 3D. Then at each timestep, we have the 3D coordinates of those features. Now, between two timesteps for which there are at least 3 of the same features triangulated, we can estimate the optimal (in terms of least square error) rigid transformation between the feature coordinates at the timesteps \citep{arun1987least}. This optimal transformation can be found by taking the vector between the centroids of the coordinates as the translation and then performing an SVD to recover the rotation. We repeat this for pairs of timesteps that share at least 3 triangulated features until we have accounted for all timesteps. If we are unable to add a timestep because it does not share sufficiently many triangulated features with another timestep, we ignore it.

Since the features we track are rigidly attached to the end-effector, the transformations between the features in 3D at two timesteps is equivalent to the transformation of the end-effector pose between those two timesteps. We can compose these transformation across timesteps to recover an estimate of the end-effector Cartesian pose at each timestep.

This procedure yields estimates of (a) the poses of the cameras; (b) the the pose of the end-effector at each timestep. Though triangulation in general can be inaccurate, we have found that it provides a sufficiently good initialization for the learning.

\subsection{Initialization by Structure from Motion}\label{sec:sfm}
Since the Triangulation method depends on two cameras, we fall back to using a Structure from Motion procedure when we only have a single camera. While, most often, Structure from Motion is applied to settings with a fixed object and a moving camera, the same method can be applied to our setting of a fixed camera and a moving object (the robot arm). We apply an incremental Structure from Motion procedure that chooses a pair of images that has feature correspondences, estimates the camera baseline, and triangulates the features to 3D \citep{schonberger2016structure}. Then, new images that observe the already triangulated features are sequentially added. As we add more images, our coverage and estimation of the structure improves. When a new feature is observed by two images, we can triangulate that new feature and add it to our model of the object's structure.

At the end of this procedure, we will have discovered a) the structure of the object; b) the poses of the centroid of that object at each timestep, which then let us find the rigid transform of the end-effector pose; and c) the poses of the cameras. Again, while this procedure may have inaccuracies, it provides a sufficiently good initialization for the learning, in practice. As is described in a later section, this procedure is quite sensitive to the two initial views that are selected. 

\subsection{Kinematic Parameter Learning}\label{sec:kinparamlearning}
We seek to learn the robot kinematic model parametrized by $R$. While the previous section learned the end-effector Cartesian pose, a node in the generative model, the kinematics learning procedure only learns the kinematics parameters of the model.

The function $D_{\text{pose}}(G, A) = \| G - A \|_F$ is the Frobenius norm between the transformation matrices $G$ and $A$, corresponding to the generated and actual end-effector Cartesian pose, respectively. We have found, in practice, that there are many Cartesian pose errors that work, including quaternion error.

The optimization is as follows:

\begin{equation}
\label{eq:losskin}
\begin{gathered}
L_{\text{kin}}(R, \mathbf{j}_{1:T}, P_{1:T})  = \\
\sum_{t=1}^{T} D_{\text{pose}}(K_{R}(\mathbf{j}_t),P_t) 
\end{gathered}
\end{equation}

\begin{equation}
\label{eq:optimkin}
\begin{gathered}
R^* = \underset{R}{\text{argmin}} \  L_{\text{kin}}(R, \mathbf{j}^o_{1:T}, P^o_{1:T})
\end{gathered}
\end{equation}
The loss in Eq. \eqref{eq:losskin} is the error between the pose generated from joint states $\mathbf{j}_{1:T}$ through the model and the poses $P_{1:T}$. In, Eq. \eqref{eq:optimkin}, the superscript $o$ on $\mathbf{j}^o_{1:T}$ and $P^o_{1:T}$ indicates that the joint states and Cartesian pose are observed. $P^o_{1:T}$ is not directly observed but we set it to be the $P_{1:T}^*$ that is output from the \nameref{sec:camstructlearning}.

Note that we assume the user will provide the number of degrees of freedom $n$ implicitly, by providing the joint states. Therefore we can include $n$ sets of DH parameters in our kinematic parameters $R$.

Unlike the learning of the camera and structure parameters, we have found that this optimization can converge from a random initialization of $R$.

\section{End-To-End Learning}
\label{sec:e2e}

Since we observe the joint states $\mathbf{j}_{1:T}$ and the feature pixel locations $I_{1:T}$, one might attempt to directly learn the function from joint states to pixel locations. However attempting to learn this from a random initialization is infeasible, as it consistently converges to sub-optimal local minima. The output of the learning discussed in \nameref{sec:camstructlearning} and \nameref{sec:kinparamlearning} provide an initialization of the parameters from which the end-to-end learning can converge to a good solution.

Additionally, the end-to-end learning is important for improving the accuracy of the learned model. While the previous stages of learning might individually converge to good solutions, composing them and using them to generate feature pixel coordinates from joint positions may not yet be accurate. This is because the end-effector Cartesian pose estimated in the first stage of learning has errors, and thus the kinematic model learned from those poses will also have errors. Optimizing all the model parameters simultaneously allows us to correct for these errors.

The end-to-end learning optimization is as follows:

\begin{equation}
\label{eq:losse2e}
\begin{gathered}
L_{\text{e2e}}(R, F, V_{1:c}, \mathbf{j}_{1:T}, I_{1:T}) =  \\ \sum_{t=1}^{T} \sum_{i=1}^{c} D_{\text{pixel}}(C_{V_i}(S_F(K_{R}(\mathbf{j}_t))),I_{ti})
\end{gathered}
\end{equation}
\begin{equation}
\label{eq:optime2e}
\begin{gathered}
R^f, F^f, V_{1:c}^f = \\ \underset{R, F, V_{1:c}}{\text{argmin}} L_{\text{e2e}}(R, F, V_{1:c}, \mathbf{j}^o_{1:T}, I^o_{1:T})
\end{gathered}
\end{equation}
We use superscript $f$ in Eq. \eqref{eq:optime2e} to denote that these are the final settings of each of these parameters. The loss in Eq. \eqref{eq:losse2e} is the error between the pixel coordinates in $I_{1:T}$ and the pixel coordinates generated through our full model from the joint states $\mathbf{j}_{1:T}$ through our full model. In Eq. \eqref{eq:optime2e}, we indicate that the joint states $\mathbf{j}^o_{1:T}$ and images $I^o_{1:T}$ are observed.

We initialize $F$ and $V_{1:c}$, to be the $F^*$ and $V_{1:c}^*$ output from the camera and structure learning in Eq. \eqref{eq:optimcamstruct}. We initialize $R$ to be the $R^*$ that is output from the kinematics learning in Eq. \eqref{eq:optimkin}.
\section{Learning from Noisy State and Images}
\label{sec:learningstate}

Here we consider the setting where we only observe the images and a noisy estimate of the robot's state (joint states in the case of joint controller and end-effector Cartesian pose in the case of a Cartesian controller).

\subsection{Noisy Cartesian Pose}
Since we are operating a Cartesian controller, we do not need to learn the robot kinematics. We only need to learn the camera and structure parameters and the true Cartesian pose of the end-effector at each timestep, correcting for the noise. Since we get a noisy estimate of the pose of the end-effector, we can bootstrap our computation of the initialization (either through triangulation or structure form motion) with these estimates. Then our optimization is as follows:

\begin{equation}
\label{eq:lossnoisycart}
\begin{gathered}
L_{\text{noisycart}}(P_{1:T},P^o_{1:T}, F, V_{1:c}, I_{1:T}) = \\ L_{\text{cam}}(P_{1:T}, F, V_{1:c}, I_{1:T}) + \lambda \sum_{i=1}^T \ D_{\text{pose}}(P_t, P^o_t)
\end{gathered}
\end{equation}

\begin{equation}
\label{eq:optimnoisycart}
\begin{gathered}
P_{1:T}^*, F^*, V_{1:c}^*= \\ \underset{P_{1:T}, F, V_{1:c}}{\text{argmin}}  L_{\text{noisycart}}(P_{1:T}, P_{1:T}^{o}, F, V_{1:c},I_{1:T}^{o})
\end{gathered}
\end{equation}
The loss in Eq. \eqref{eq:lossnoisycart} extends the loss in Eq. \eqref{eq:losscamstruct} with an the additional term that penalizes the learned pose from differing from the observed pose. In the additional term, $P^o_t$ is the noisy estimate of the end-effector pose that we observe. $\lambda$ is a weight that captures how much we trust the noisy pose estimates. It will vary inversely proportionally with the variance of the noise in the observed Cartesian pose.

In both the above equations, we have both $P_{1:T}$ and $P^o_{1:T}$. $P_{1:T}$ are the Cartesian poses that we learn, so they are free parameter in the optimization process. $P^o_{1:T}$ is the Cartesian pose that we observe, which is noisy.

\subsection{Noisy Joint States}
We observe noisy estimates of the joint states at each timestep. We perform the \nameref{sec:camstructlearning} and \nameref{sec:kinparamlearning} using the noisy joint state estimates, which will produce an inaccurate model but serves as a good initialization. The function $D_{\text{joint}}(\mathbf{j}_1, \mathbf{j}_2) = \| \mathbf{j}_1 - \mathbf{j}_2 \|_2^2$ is an error between joint states. The optimization to correct the inaccurate model is:

\begin{equation}
\label{eq:lossnoisyjoints}
\begin{gathered}
L_{\text{noisyjoint}}(\mathbf{j}_{1:T}, \mathbf{j}^o_{1:T}, R, F, V_{1:c}, I_{1:T}) = \\
L_{\text{e2e}}(R, F, V_{1:c}, \mathbf{j}_{1:T}, I_{1:T}) + \lambda \sum_{i=1}^T \ D_{\text{joint}}(\mathbf{j}_t, \mathbf{j}^o_t)
\end{gathered}
\end{equation}

\begin{equation}
\label{eq:optimnoisyjoints}
\begin{gathered}
\mathbf{j}_{1:T}^*, R^*, F^*, V_{1:c}^* = \\
\underset{\mathbf{j}_{1:T}, R, F, V_{1:c}}{\text{argmin}} L_{\text{noisyjoint}}( \mathbf{j}_{1:T},\mathbf{j}^o_{1:T}, R, F, V_{1:c},  I_{1:T}^o)
\end{gathered}
\end{equation}
The loss in Eq. \eqref{eq:lossnoisyjoints} extends the end-to-end loss in Eq. \eqref{eq:losse2e} with an  additional term that penalizes the learned joint states from differing from the observed joint states. In the additional term, $\mathbf{j}^o_t$ is the noisy estimate of the joint states that we observe. $\lambda$ is a weight that captures how much we trust the noisy joint estimates. Again, it will vary inversely proportionally with the variance of the noise in the observed joint states. 

$\mathbf{j}_{1:T}$ and $\mathbf{j}^o_{1:T}$ appear in both the above equations. $\mathbf{j}_{1:T}$ are the joint states that we learn. $\mathbf{j}^o_{1:T}$ are the noisy observations of the joint states.

Note that both the above settings contain the noiseless state setting as a special case. As the noise goes to zero, we can increase $\lambda \xrightarrow{} \infty$ and recover the exact learning equations of the noiseless case as shown in  \nameref{sec:camstructlearning} and \nameref{sec:e2e}, respectively. 
\section{Learning from Noisy Actions and Images}
\label{sec:learningactions}

Here we consider the setting where we cannot read the state (joint states or pose of the end effector) of the robot, but only the actions that we send to a noisy Cartesian or joint controller and the image features. This setting is both theoretically and practically interesting as it allows us to demonstrate how uncertainty can be handled in our model in order to improve control of robots with inaccurate controllers. 

\subsection{Noisy Cartesian Controller}

In the case of a Cartesian controller, the actions we send are desired transformations of the end-effector. We first simply assume the actions are noiseless and compose them to get an estimate of the end-effector pose at each frame. Then independently of this, we learn the camera and structure parameters and end-effector pose at each timestep as is done in \nameref{sec:camstructlearning} and \nameref{sec:kinparamlearning}. Now we have two estimates of the end effector pose that are within an arbitrary scaling and transformation of one another. We run an optimization to find a scaling and transformation that minimizes the pose error between these independent estimates of poses. Then we apply this scaling and transformation to the learned poses, structure parameters, and camera extrinsics. Now we have end-effector poses that are more consistent with the actions as well as the same reconstruction error from the independent learning of camera and feature parameters. In the final stage, we run the following optimization:

\begin{equation}
\label{eq:losscartaction}
\begin{gathered}
L_{\text{cartaction}}(P_{1:T}, F, V_{1:c}, I_{1:T}, A_{1:T}) = \\ L_{\text{cam}}(P_{1:T}, F, V_{1:c}, I_{1:T}) +\\ \lambda \sum_{t=1}^{T-1} \ D_{\text{pose}}(P_{t+1}P_t^{-1}, A_t)
\end{gathered}
\end{equation}
\begin{equation}
\label{eq:optimcartaction}
\begin{gathered}
P_{1:T}^*, F^*, V_{1:c}^*= \\ \underset{P_{1:T}, F, V_{1:c}}{\text{argmin}}  L_{\text{cartaction}}(P_{1:T}, F, V_{1:c},I_{1:T}^{o}, A_{1:T}^{o})
\end{gathered}
\end{equation}
The loss in Eq. \eqref{eq:losscartaction} modifies the loss in Eq. \eqref{eq:lossnoisycart} with an additional term that penalize consecutive end-effector poses differing from the commanded transformation $A_t$ at that timestep. Only the actions $A^o_{1:T}$ and pixel coordinates $I_{1:T}^{o}$ are observed.

\subsection{Noisy Joint Controller}
In the case of a joint controller, the actions we send are desired deltas of the joint angles. We first learn the camera and feature parameters as is done in \nameref{sec:camstructlearning}. Then, we learn the kinematics parameters as is done in \nameref{sec:kinparamlearning}, however instead of using the ground truth joint angles, which are unavailable, we assume the joint deltas are noiseless and compose them to get joint angles at each step. Then we run the end-to-end learning from Section \nameref{sec:e2e}. In the final stage, we make the joint angles a free parameter and run the following optimization:

\begin{equation}
\label{eq:lossjointaction}
\begin{gathered}
L_{\text{jointaction}}(\mathbf{j}_{1:T}, R, F, V_{1:c}, I_{1:T}, \mathbf{\hat{a}}_{1:T}) = \\
L_{\text{e2e}}(R, F, V_{1:c}, \mathbf{j}_{1:T}, I_{1:T}) + \\ \lambda \sum_{t=1}^{T-1} \ D_{\text{joint}}(\mathbf{j}_{t+1} - \mathbf{j}_t, \mathbf{\hat{a}}_t)
\end{gathered}
\end{equation}

\begin{equation}
\label{eq:optimjointaction}
\begin{gathered}
\mathbf{j}_{1:T}^*, R^*, F^*, V_{1:c}^* = \\
\underset{\mathbf{j}_{1:T}, R, F, V_{1:c}}{\text{argmin}} L_{\text{noisyjoint}}( \mathbf{j}_{1:T}, R, F, V_{1:c},  I_{1:T}^o, \mathbf{a}_{1:T}^o)
\end{gathered}
\end{equation}
The loss in Eq. \eqref{eq:lossjointaction} modifies the loss in Eq. \eqref{eq:lossnoisyjoints} by having the additional term penalize consecutive joint states differing from the commanded delta of the joint states $\mathbf{a}_t$ at that timestep. Only the $\mathbf{a}_{1:T}^o$ and pixel coordinates $ I_{1:T}^o$ are observed.
\section{Non-physical Parameters}
\label{sec:nonphys}

It is important to note that the parameters of our model do not necessarily correspond to the true physical setup. While a properly learned model can accurately generate image pixel locations from joint states, the latent variables are non-physical. There is a degeneracy up to an arbitrary scale and rigid transformation. For example, if we scale up, rotate, and translate the space, we can produce an equivalent model in terms of the generation of features' pixel coordinates from joint states.

To resolve this degeneracy, we would need information from the actual physical setup. If we provided the actual scale of the structure or the camera extrinsic parameters we could correct the scale ambiguity. Additionally if we corrected the base frame Cartesian pose or gave the true camera extrinsics, we could correct the transform ambiguity. However as our demonstrations show, physicality is not necessary in most visual control settings.

\section{Optimization}
\label{sec:optim}
The optimizer we use for learning in this model is the quasi-Newton method limited-memory BFGS \citep{liu1989limited} which approximates the Hessian matrix to make updates. All of the forward computations in our model are differentiable and the gradients of all parameters can be computed analytically. This allows for fast and accurate updates, which improves the overall efficiency and performance of the learning procedure. We use Pytorch for automatic differentiation and optimization \citep{paszke2017automatic}.

Additionally, we have found that learning using mini-batches of the dataset can help to escape local minima. We implemented a version of L-BFGS that first learns on random minibatches of the data before a final period of learning with all the data.
\section{Inference}
\label{sec:inference}

Using our learned model, we would like to be able to efficiently and accurately control the robot using visual feedback. Specifically, given target locations of the features in the image, we want to be able to command the robot to match those targets. This will involve inferring the end effector pose and corresponding joint states that would place the features at the target positions. Since we have the learned the structure of the features in the end-effector frame as well as the camera parameters, when given target pixel locations of the features, we can infer the end-effector Cartesian pose with the Perspective n Point algorithm \citep{lepetit2009epnp}. Then, with that end-effector pose and the kinematic parameters of the robot, we can solve the standard inverse kinematics (IK) problem to recover the joint states of the robot \citep{beeson2015trac}. Since the map from joint states to end effector pose is non-injective, the IK procedure can return multiple solutions for the joint states. When we are operating the robot by directly commanding joint states, this does not pose a problem because we can command the robot to any one of the joint state solutions and we will achieve the feature locations in image space. However when we can only command desired deltas of the joint states, if we infer our current joint states incorrectly, then the joint deltas we will send to the robot will not result in the desired motion. We can solve this issue in two ways. The first is by tracking features on earlier links of the robot (those before the end effector). Doing so will allow us to infer those earlier joint's states and help us filter the joint states output by the IK. Another method to solve this is by finding the joint state that is most consistent with the actions that we have taken. Since this does not require tracking of any additional markers, we chose to use this method.

Now given the end effector pose and joint states of our current and target locations, we can properly command the robot. If we are commanding the robot with joint states directly, then we will simply move to the joint states inferred by our model. If we can only command the robot using deltas in joint or Cartesian space then we will compute the delta between our inferred current state and inferred target state and send this as a command. With an ideal model and system, we will match the target features accurately in one motion. In our experiments, we will see that when using an accurate robot this is possible. However with noisy controllers, this procedure might need to be repeated several times before the target features are matched.

\section{Model Properties}
\label{sec:modelproperties}

In this section we review specific properties of our model. We discuss the importance of the end-to-end learning, the speed of inference, and the effectiveness of the complete learning procedure.

\subsection{End-to-End Learning}
We have found that end-to-end learning of all the model parameters can greatly improve the model's performance. Table \ref{tab:endtoend} shows the reconstruction errors achieved with and without learning end to end. ``Before end-to-end'' means the camera and structure parameters were learned first and then the kinematic parameters. ``After end-to-end'' initializes from there and learns all parameters simultaneously.

\begin{table}[ht]
\small\sf\centering
\caption{Loss reduction obtained through end-to-end learning as shown in eq. \eqref{eq:losse2e} and eq. \eqref{eq:optime2e} }
\begin{tabular}{lll}
\toprule
- & Simulation & Real\\
\midrule
Before end-to-end & 414.21 & 324.16\\
After end-to-end & \textbf{0.47} & \textbf{0.34} \\
\bottomrule
\end{tabular}\\[10pt]
\label{tab:endtoend}
\end{table}

However, when learning the model parameters, it is insufficient to learn all of them simultaneously from random initialization. Instead we found that learning subsets of the parameters in stages, with each stage using the result of the previous one as initialization, had the best performance. In Table \ref{tab:stages}, we compare the average convergence in three settings: (1) learning all parameters simultaneously (2) learning camera  and feature parameters and then all parameters simultaneously (3) learning camera and feature parameters, then kinematic parameters, and finally all simultaneously, which performed best. In each of these we use 600 training iterations.

\begin{table}[ht]
\small\sf\centering
\caption{Stages of optimization}
\begin{tabular}{lll}
\toprule
- & Simulation & Real\\
\midrule
All & 549.4 & 173.2 \\
Cam + Struct, All &  27077.9 & 8985.9 \\
Cam + Struct, Kin., All & \textbf{0.47} & \textbf{0.34} \\
\bottomrule
\end{tabular}\\[10pt]
\label{tab:stages}
\end{table}

End-to-end learning is practically important because when independently calibrating kinematics and cameras, errors in one calibration can produce errors in the other. For example, with an inaccurate kinematics model, reprojection error from a calibration procedure might be significantly higher. We have found that errors in kinematics calibration can lead to increased reprojection errors in camera calibration. 

\subsection{Fast Inference}
Recent work in uncalibrated servoing has used neural networks. An advantage of these architectures is its speed, which allows for real-time servoing. In Table \ref{tab:speed}, we show the speeds of various learning and inference queries. The results are for a setup with 2 cameras, 12 features, and a 6 joint robot.

\begin{table}[ht]
\small\sf\centering
\caption{Speed of queries}
\begin{tabular}{llll}
\toprule
Query & Time (ms)\\
\midrule
Forward Kinematics & 1.27 (0.01) \\
Forward Camera/Feature & 1.43 (0.01) \\
Forward End-to-End & 2.58 (0.01) \\
Loss (Pixel) & 0.041 (0.003) \\
Loss (Cartesian Pose) & 0.0785 (0.004) \\
Gradient End-to-End & 16.1 (0.81) \\
 Gradient Kinematics & 12.6 (0.45) \\ 
 Gradient Camera/Feature & 9.12 (0.37) \\ 
Infer Pose from Image & 0.511 (0.185) \\
Infer Joints from Pose & 101 (1.1) \\ 
Infer Joints from Image & 102 (1.1) \\ 
 
\bottomrule
\end{tabular}\\[10pt]
\label{tab:speed}
\end{table}

Compared to the other operations in our model, inferences of the joints from the image are significantly slower. This is because it is an optimization process that requires multiple evaluations of a function and its gradient.

While we do not see this as a limitation for real-time servoing, if an application required a high-frequency control signal, we would simply use an amortized inference technique. For instance, we can simply learn a neural network as a backward model. This backward model can be trained offline without further data capture since an arbitrary number of examples can be produced by the forward model we have already learned.

\subsection{Training Procedure}
Our model is trained in the following 3 stages given the initialization:

\begin{enumerate}
  \item Learn camera parameters, structure parameters, and end-effector pose given the feature pixel locations. (200 iterations)
  \item Learn kinematics parameters given joint states and end-effector pose (200 iterations with minibatch mode)
  \item Learn all parameters given joint states and feature pixel locations (200 iterations with minibatch mode)
\end{enumerate}

Using this procedure, training gives consistent converge to good solutions. In Table \ref{tab:kfold} we run a $k$-fold cross validation for different values of $k$. The dataset consists of 200 samples of joint states and corresponding image feature locations. We show consistently converge to a solution that generalizes well to test data. 

\begin{table}[ht]
\small\sf\centering
\caption{k-fold Cross Validation}
\begin{tabular}{cccc}
\toprule
k & Train/Val. Size & \multicolumn{2}{c}{Average Pixel Norm on Val. Set} \\
\midrule
&&Simulation&Real\\
\midrule
2 & 100/100 & 0.5600 (0.0205) & 0.1051 (0.0035) \\
3 & 132/66 & 0.5502 (0.0417)& 0.1248 (0.0197) \\
4 & 150/50 & 0.5577 (0.0663) & 0.1130 (0.0059) \\
5 & 160/40 & 0.6368 (0.2268)& 0.1154 (0.0083)\\
6 & 165/33 & 0.5445 (0.0667)& 0.1181 (0.0090) \\
\bottomrule
\end{tabular}\\[10pt]
\label{tab:kfold}
\end{table}

\subsection{Sample Efficiency}
However though the model's consistent convergence is important, we are ultimately interested in its performance in the the low-data regime. To evaluate this, we hold-out 100 samples and take smaller subsets of the remaining 100 samples for training. Table \ref{tab:sampleeff} shows the performance on test data for each size training set.

\begin{table}[ht]
\small\sf\centering
\caption{Performance with few samples}
\begin{tabular}{ccc}
\toprule
Training Samples &  \multicolumn{2}{c}{Average Pixel Norm on Val. Set} \\
\midrule
&Simulation&Real\\
\midrule
30 & 49.3582 (43.5229) & 0.1367 (0.0032) \\
40  & 3.2066 (2.3041) & 0.1331 (0.0081) \\
50  & 0.6235 (0.0143)& 0.1704 (0.0416) \\
60  & 0.6168 (0.0082) & 0.1194 (0.0021)  \\
70  & 0.5991 (0.0082) & 0.1170 (0.0023) \\
\bottomrule
\end{tabular}\\[10pt]
\label{tab:sampleeff}
\end{table}

\begin{figure}\centering\includegraphics[width=\linewidth]{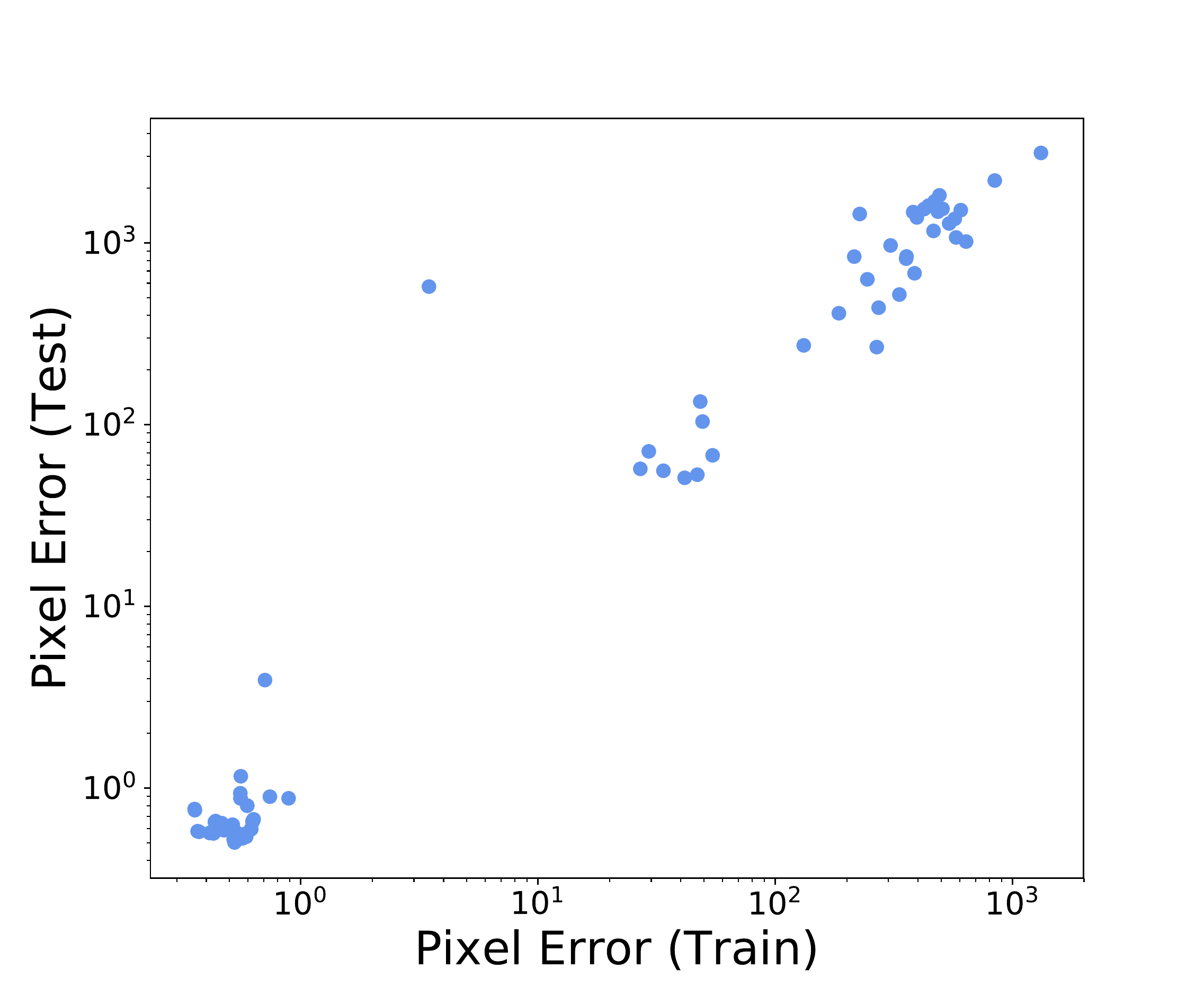}\caption{Train vs. Test Pixel Error}
\label{fig:scat}
\end{figure}

We have found that our training performance correlates strongly with our test performance. Each data point in Figure \ref{fig:scat} corresponds to a training period with 50 randomly selected training samples and 100 test samples. It shows that when training error converges to sub pixel values, the test error does as well.

\section{Model Flexibility}
\label{sec:flexibility}

The modularity of our model enables us to swap out, add, and remove components easily. We discuss how this is useful for each of the classes of parameters in the model: cameras, structure, and  kinematics.

\subsection{Cameras}
Users may want to modify the number and/or position of cameras in their setup. This might be needed when trying to do fine manipulation tasks in which having a camera close to the robot end-effector can improve accuracy. We show that we can relearn cameras and make them usable for guiding motions with very few iterations.

\begin{figure}\centering\includegraphics[width=\linewidth]{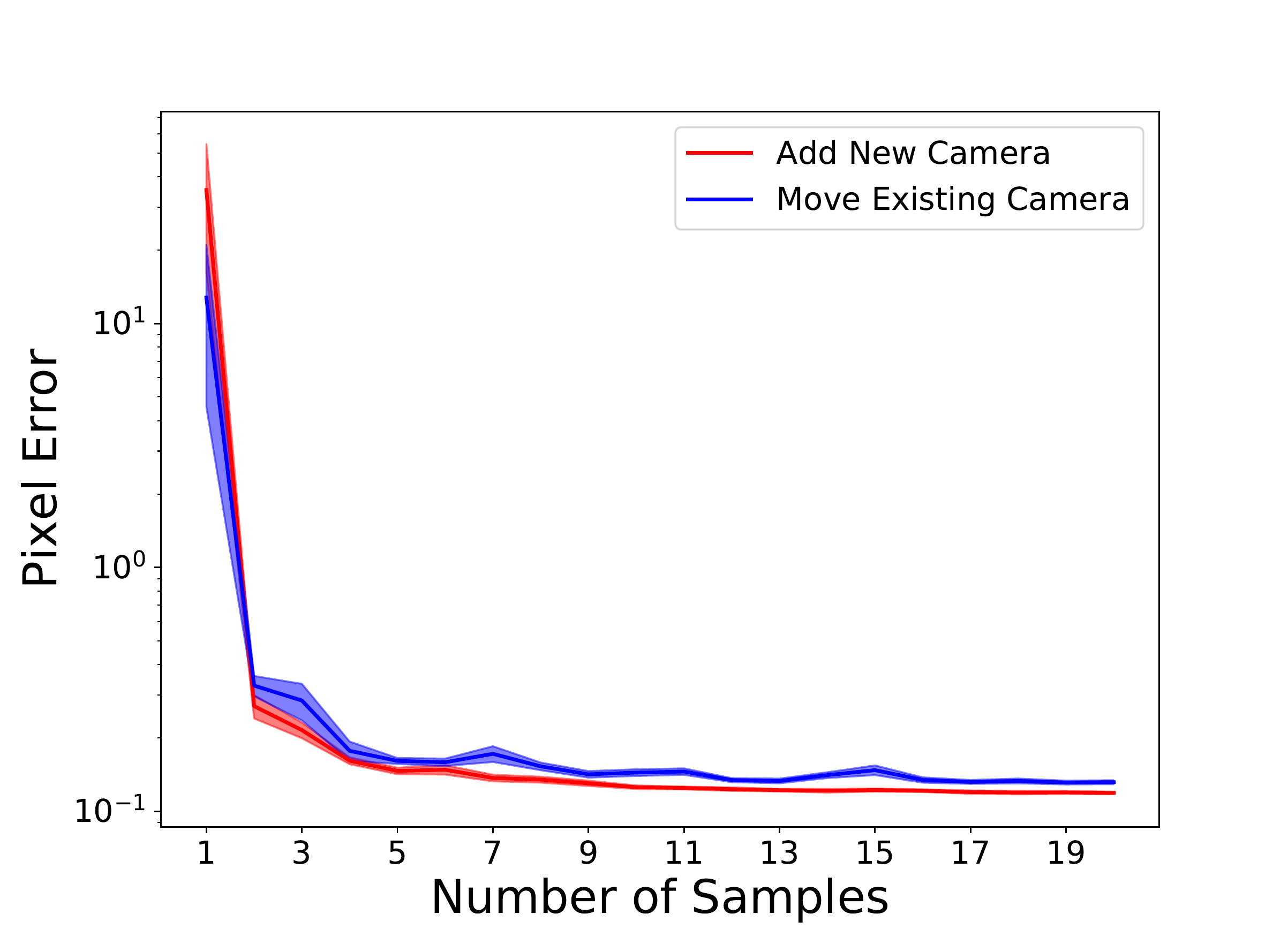}\caption{Samples to learn new camera parameters}\label{fig:newcam}\end{figure}

Figure \ref{fig:newcam} shows the pixel error on a test set as we increase the number of samples. The camera parameters are being learned from random initialization. While the learning does not converge given only 2 samples, it is already sufficiently accurate to begin servoing. The plot shows both the case where add a third camera and the case where we have two cameras and we move one of them.

In addition, our method does not depend on having multiple cameras. Our learning procedure is feasible with a single camera using the initialization described in \nameref{sec:sfm}. Since this method of computing an initialization is sensitive to the specific views that are chosen, we run our learning with many more restarts than are used for the two-camera setting. We compared the pixel errors on a test data set when training with a single camera. When we use two cameras to compute an initialization (using triangulation) and then learn using information from only a single camera, we get a pixel error of $0.541 \ (0.187)$. When we use only a single camera to compute to compute an initialization (using structure from motion), we get a pixel error of $1.231 \ (0.705)$. This is done over 10 trials and with data collection from the real robot. While the performance of the learning degrades when using only a single camera, it still allows for accurate positioning. In practice though, we use two cameras because it is more accurate and faster to train, since it requires less restarts of the optimization.

Inference in our model can also be done with a single camera. Table \ref{tab:singlecam} compares the joint and reprojection errors when doing inference with two cameras vs. with only one, after training a model with two cameras.

\begin{table}[h]
\small\sf\centering
\caption{Single Camera Joint Inference Accuracy and Reprojection Error}
\begin{tabular}{ccc}
\toprule
Error & Simulation & Real \\
\midrule
Both - Joint & 0.0536 (0.0134) & 0.1167 (0.1167) \\
Single - Joint& 0.0519  (0.0095) & 0.1544 (0.0243) \\
Both - Reprojection & 1.1441 (0.1450) & 0.7661 (0.0997) \\
Single - Reprojection& 1.6608 (0.2262) & 1.0329 (1.0329)\\
\bottomrule
\end{tabular}\\[10pt]
\label{tab:singlecam}
\end{table}

\subsection{Structure}
An important use of robots is in performing manipulation tasks. While the above sections have assumed that the features we track are fixed to the robot arm, in practical use cases these features may shift. For example, if we grasp an object and are uncertain about its position in the gripper, we may want to use visual feedback to infer the new structure in order to begin accurately manipulating that object. In our model, adding new features, potentially from a new object, is simgs/traightforward and learning the parameters can be done in a sample efficient manner online. We demonstrate this by attaching new markers to the robot gripper and removing the previous markers. We show that within a few data samples, we can accurately position those new markers.

\begin{figure}\centering\includegraphics[width=\linewidth]{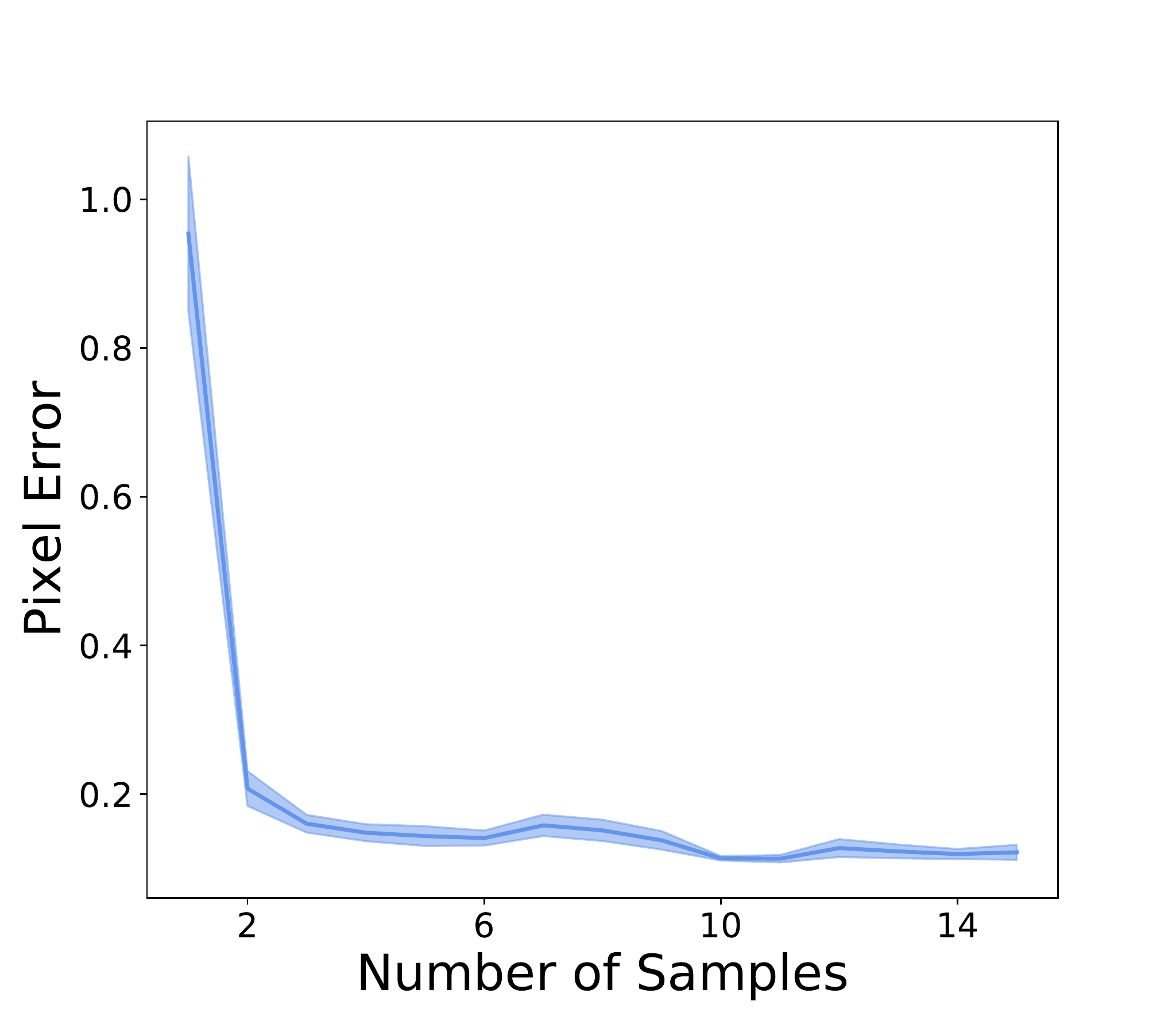}\caption{Samples to learn new structure parameters}
\label{fig:newstr}
\end{figure}

Figure \ref{fig:newstr} shows the reprojection error when we learn the missing structure parameters as a function of the number of samples. Again, the structure parameters are initialized randomly. With 2 samples, the model is very close to maximum precision. One way to explain this efficiency is to think that if we have two cameras properly learned in the model, the new features' pixel locations can be triangulated to 3D. As we get more views, the effect of the noise from the triangulation reduces. Note that we do not explicitly do this triangulation, rather it is done implicitly through the optimization of the structure coordinates.

\subsection{Kinematics}
\begin{figure}\centering\includegraphics[width=\linewidth]{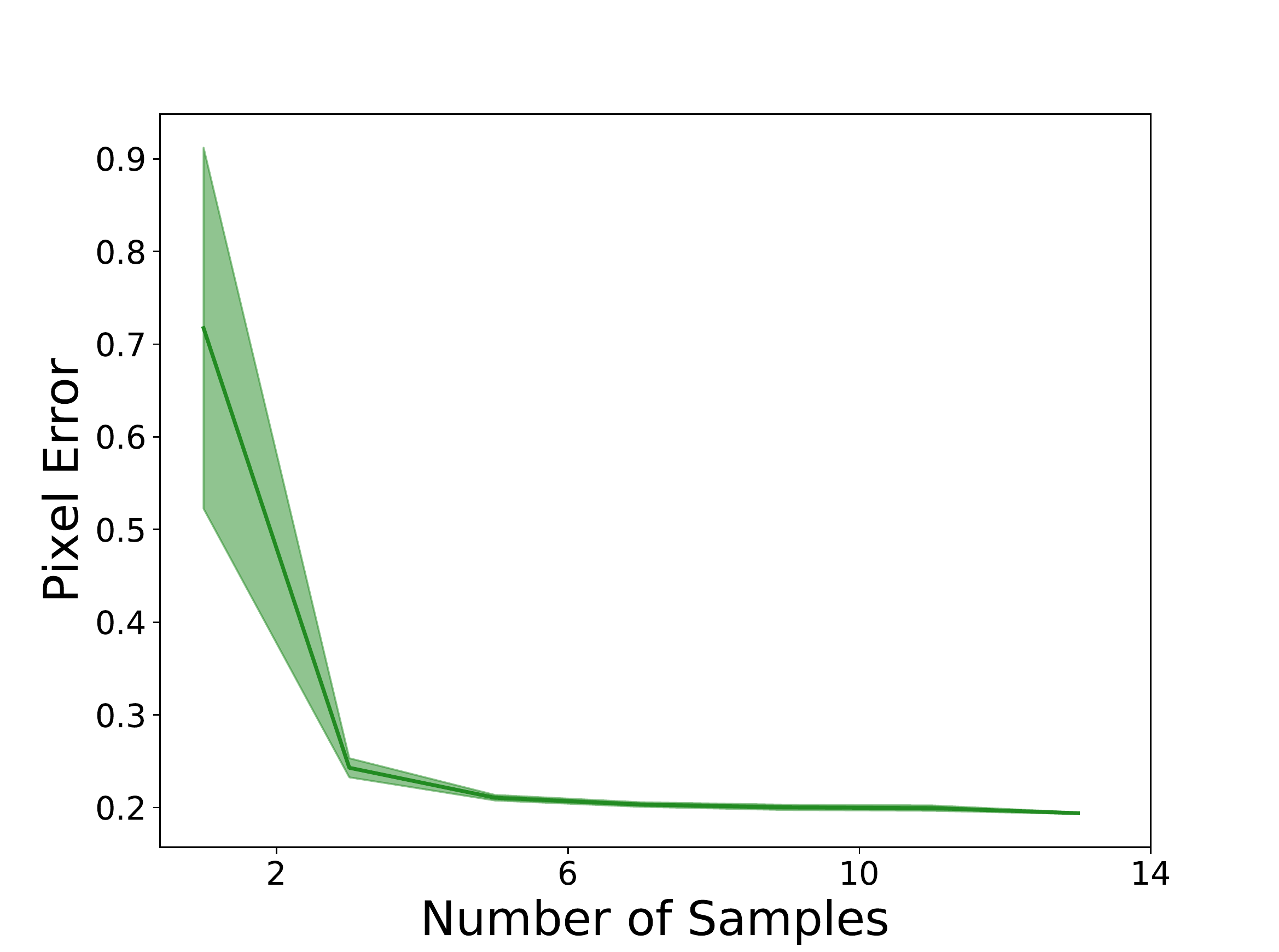}\caption{Samples to learn new base frame parameters}
\label{fig:newbase}
\end{figure}

The kinematic parameters might change for a variety of reasons. For example, if the robot is not fixed to the ground, motions of the arm can cause the base of the robot to shift. In Figure \ref{fig:newbase}, we show how the new base frame of the robot can be learned with very few samples.

\begin{figure}\centering\includegraphics[width=\linewidth]{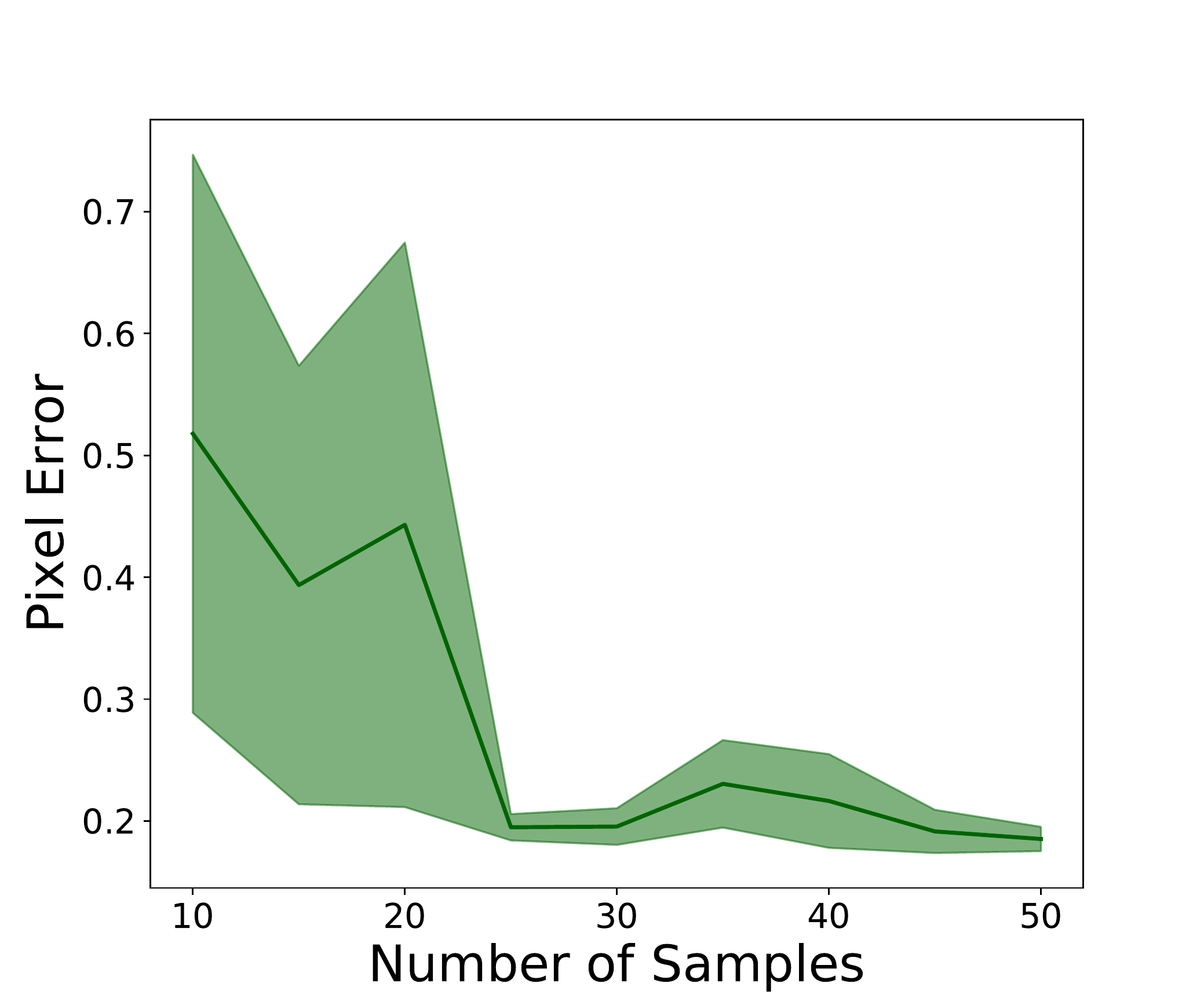}\caption{Samples to learn new robot kinematic parameters}
\label{fig:newrobot}
\end{figure}

The kinematic calibration of the arm could also change if we attach new grippers, force-torque sensors, or modify the arm. While we have not collected data where we modify the arm, we can get a sense of whether relearning the kinematics is possible by randomly re-initializing the kinematic parameters in our learned model and attempting to relearn them. Figure \ref{fig:newrobot} shows the convergence of this learning. Compared to camera and structure parameters, kinematic parameters require significantly more samples to consistently converge.

\section{Online Learning and Change-Point Detection}
\label{sec:online}

The above sections show that learning in our model can be done in a sample efficient manner. In addition, after having properly learned the model, relearning or adapting parameters as they change can also be done quickly. This suggests an online learning formulation for our model.

Online learning can be done by maintaining a buffer of recorded observations and taking gradients on these samples to update the parameters. In theory, these gradients should be approximately zero if the setup has not changed and the model is already converged. However, due to noise in the sensor readings they may be nonzero, so we use a threshold on the gradients to ensure stability.

The above works when the changes to the parameters are small, for example when the in-hand object shifts or the camera intrinsics change. However, when there are large changes to the system, there may be a need for significant changes to the model parameters. In this situation, simply updating the parameters using gradients and still using the incorrect model for motion will be problematic. By doing inference in our model, we can detect these large changes to the system and modify the model to keep it usable.

For example, if a camera is shifted, Table \ref{tab:gradientspike} shows that the norm of the gradients will spike as the buffer fills up with observations from the newly configured setup.

\begin{table}[h]
\small\sf\centering
\caption{Gradient Norm when camera is shifted}
\begin{tabular}{cc}
\toprule
Buffer (Samples Before/After Shift) &  Norm of Gradients \\
\midrule
50/0 & 40.618 (01.986) \\
49/1 & 945.384 (50.604) \\
48/2 & 1788.127 (73.690) \\
47/3 & 2666.490 (85.969) \\
46/4 & 3404.664 (93.083) \\
45/5 & 4287.491 (99.077) \\
44/6 & 5127.693 (111.578) \\
43/7 & 5930.182 (128.776) \\
42/8 & 6762.114 (125.875) \\
41/9 & 7630.604 (119.384) \\
40/10 & 8403.353 (121.102) \\

\bottomrule
\end{tabular}\\[10pt]
\label{tab:gradientspike}
\end{table}

After detecting this spike, we can identify the camera that is producing results inconsistent with what our model predicts. Then we ignore this camera until we accumulate sufficient samples to relearn its parameters. The previous section shows that very few samples are needed to do this.

\section{Fast Servoing and Comparison with VISP}
\label{sec:fastservoing}

Visual servoing is the technique of using visual feedback to control a robot. Most often, visual servoing tracks certain features in the images and controls the robot to optimize an objective defined on those features. This might be to match some target locations of those features or to orient them in a certain configuration. Visual servoing methods like IBVS estimate the image space Jacobian, which defines how changes to the joints will change the image, and the control law takes steps to maximize the objective function.

We compare our learned visual control to VISP, the standard implementation of IBVS, which uses the control law shown in eq. \eqref{eq:visp} for servoing. The robot Jacobian expressed in the end-effector frame and camera extrinsics calibration are given and the interaction matrix is estimated online. We must also provide the pixel and depth locations of the features as they are currently and as they are desired to be.

\begin{figure}\centering\includegraphics[width=\linewidth]{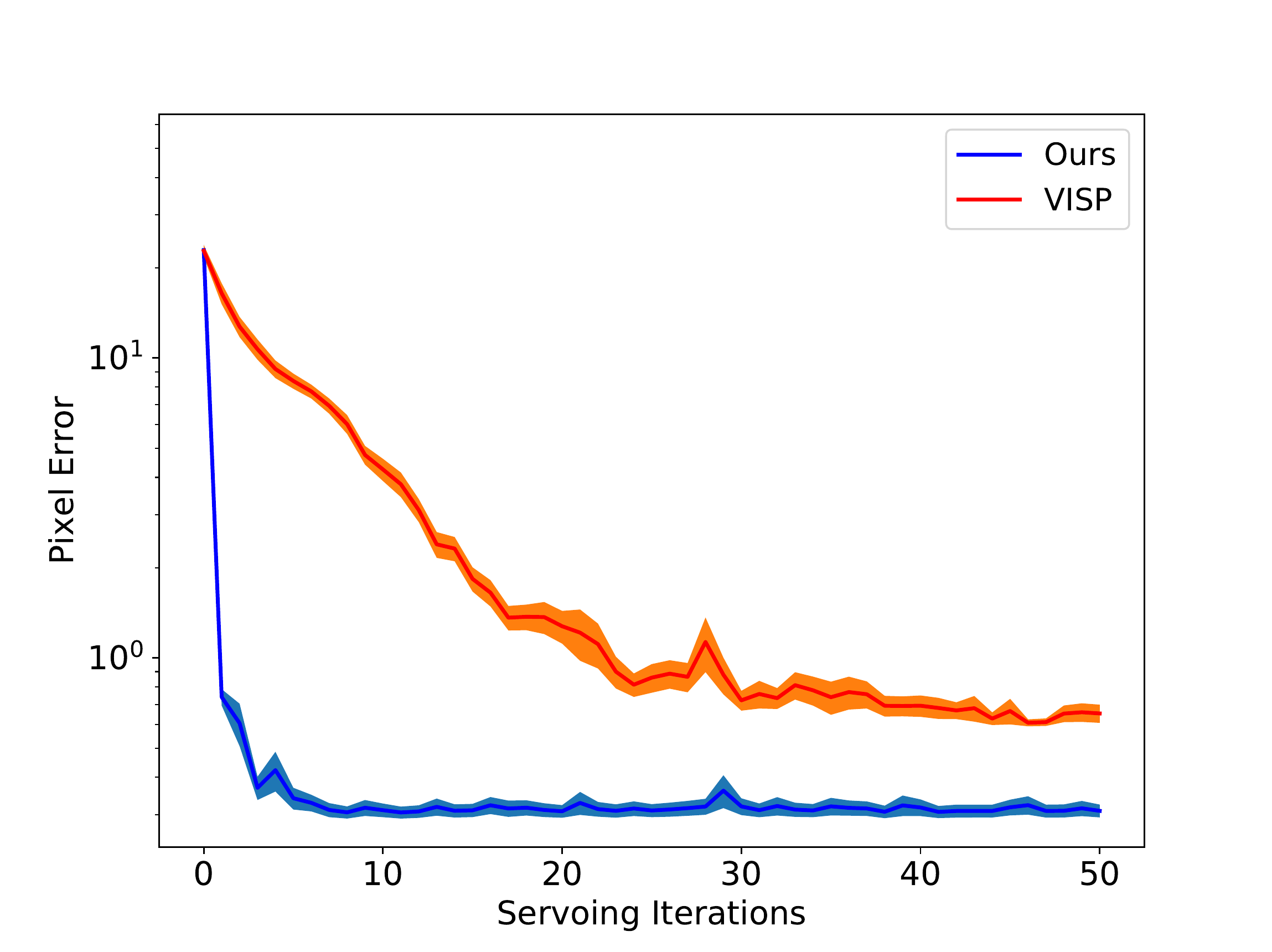}\caption{Pixel error vs. number of iterations of servoing}\label{fig:visp}\end{figure}

To compare with our model, we randomly select a target configuration of the arm and capture the features' pixel coordinates. Then we move the arm elsewhere and allow both models to servo given only the features' pixel locations at the target configuration. We record the pixel error at each step of the servoing. We repeat this procedure for 100 different target configurations. Figure \ref{fig:visp} shows that our model outperforms VISP both by doing more accurate servoing and in significantly fewer iterations.

\begin{figure}\centering\includegraphics[width=\linewidth]{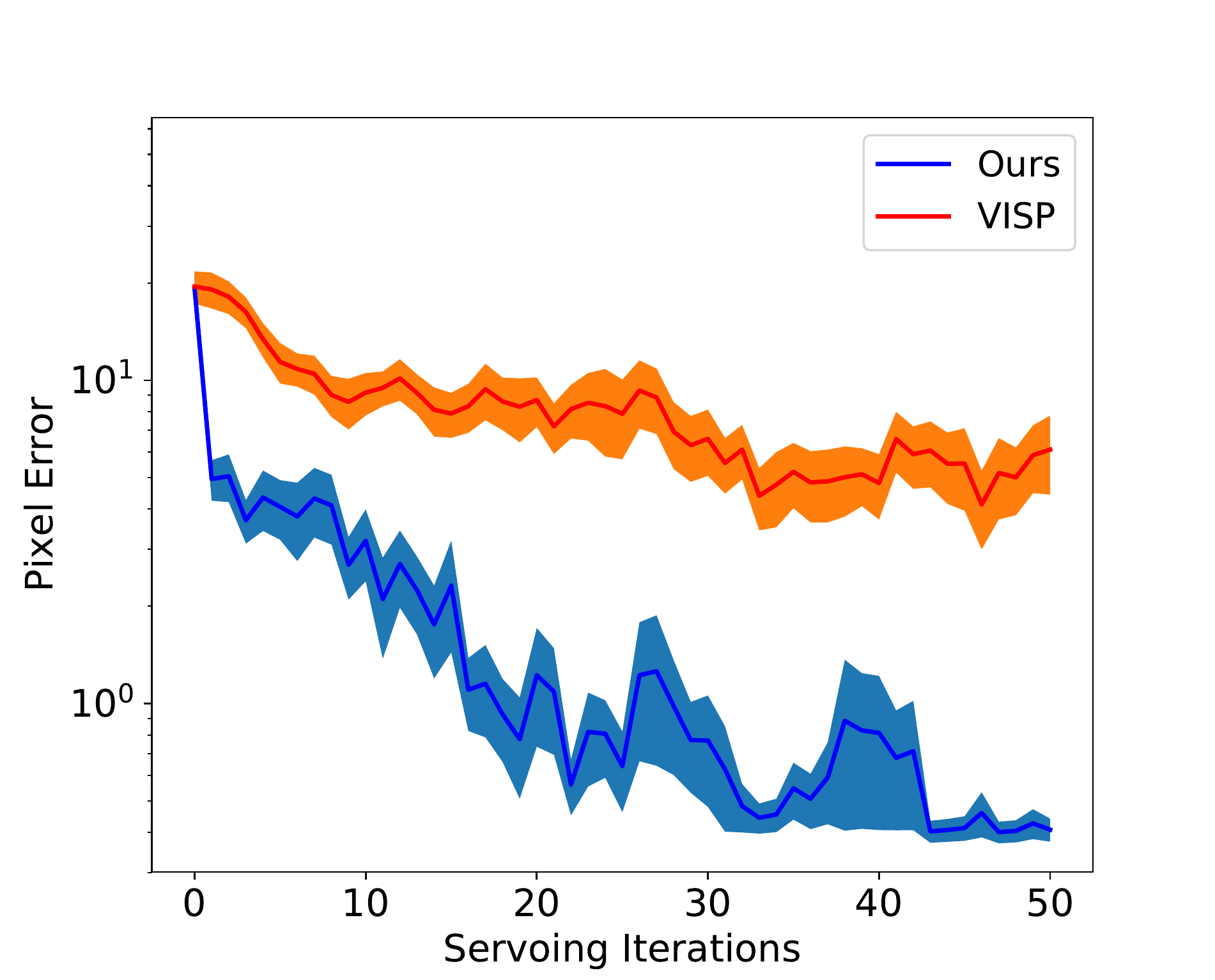}\caption{Pixel error vs. number of iterations of servoing with noisy joint states and controller}\label{fig:vispnoisy}\end{figure}

In Figure \ref{fig:vispnoisy}, we run the same evaluation but introduce noise into both the joint states and the actions sent to the controller. We add Gaussian noise with standard deviation $0.02$. Our method is trained on data with noisy joint estimates. VISP is provided the Jacobian at the noisy joint state. Our method still outperforms VISP. Since our method is able to directly infer the joints, it is more resilient to the noisy joint states that are observed.

\subsection{Calibration}
VISP requires a calibrated robot and camera. Calibration of the robot kinematics often requires a precisely manufactured device with which a calibration routine is run. Calibration of the cameras requires a calibration checkerboard. Each of these procedures can have errors and the errors in the kinematic calibration can lead to inaccurate calibration of the cameras. We have observed on the UR5 robot that small deviations from nominal kinematic parameters can result in an increase in camera reprojection errors of approximately 2 pixels. While our model does require data to be collected, it does not need any explicit calibration because the model parameters can be learned. 

\subsection{Cameras}

VISP also requires a stereo camera setup and proper intrinsic calibration since we must triangulate the points to 3D in order to recover the distance of each feature from the camera origin. Our method does not require the two camera and can fit the intrinsic parameters.

\subsection{Global}

Since our method learns a full kinematics model, the structure of the object, and the camera projections, we have a global model. This means that we do not need to use a image-joint Jacobian and rather, can directly compute a target joint configuration given an image. This allows us to make larger motions and reach the target location in significantly fewer servoing steps.

We use the our model for motion by inferring the joint states corresponding to the target location, given the target pixel locations of the features. We then infer the joint states given the current pixel locations and command the difference between these to the robot. In a perfect model, this should take a single step. However because of inaccuracies in the model, repeating this procedure will improve convergence.

Since we are directly computing the delta in joint space, there is no need to have a $\gamma$ term that controls the step size taken using the image Jacobian. With our method, we compute a delta between our current and desired state, rather than using gradients. Therefore, we do not need to specify a step size. Our method has no hyperparameters to tune.

\section{Noisy Control}
\label{sec:noisycontrol}

Incorporating visual feedback in the control loop is necessary for accurately controlling noisy robots. In this section we consider two settings of controlling the robot through a noisy controller (one that inaccurately executes the desired action). In the first, the joints are observed but there is noise in the reading. In the second, we don't observe the joints/cartesian pose at all. This is the most practically interesting setting because it shows that by only commanding the robot and observing it visually, we can learn to accurately control the robot. 

\subsection{Joints Observed}
We first show that if we do not account for the noise in the state estimates the learned model will be inaccurate. Table \ref{tab:jointnoise} compares the errors of models that do and do not learn corrections to the noisy joint state readings. The models that account for noise consistently outperform those that do not. The $\lambda$ parameter is selected with a validation set. We train on noisy data and use noiseless data at test time to evaluate the accuracy of the model.

\begin{table}[h]
\small\sf\centering
\caption{Error when learning accounting for noise}
\begin{tabular}{cccc}
\toprule
$\sigma$ & Model Noise & \multicolumn{2}{c}{Error}\\
\midrule
&& Train & Test \\
0.01 & No & 1.484 (0.061)& 1.715 (0.086)\\
0.01 & Yes & 1.218 (0.088) & 1.394 (0.109)\\
0.02 & No & 3.037 (0.153)& 3.393 (0.211)\\
0.02 & Yes & 2.133 (0.272)&  2.298 (0.274)\\
0.03 & No & 5.185 (0.347)& 6.077 (0.453)\\
0.03 & Yes &  3.859 (0.532) & 4.434 (0.635)\\
\bottomrule
\end{tabular}\\[10pt]
\label{tab:jointnoise}
\end{table}

While \ref{tab:jointnoise} shows the results for noisy joint states, the same trend is observed when we get a noisy measurement of the Cartesian pose of the end-effector and the learning is done as shown in Eq. \eqref{eq:optimnoisycart}.

\subsection{State Unobserved}

When the state is unobserved, the only observed information is the actions we send to the controller and the images. Since the actions are not necessarily executed accurately, by composing them together we get an estimate of the state that increases in variance over time. This is different than the above section (noisy observed state) in which the variance of the noise is the same at each time step. We demonstrate that we are still able to learn and then control the robot.

\begin{figure}\centering\includegraphics[width=\linewidth]{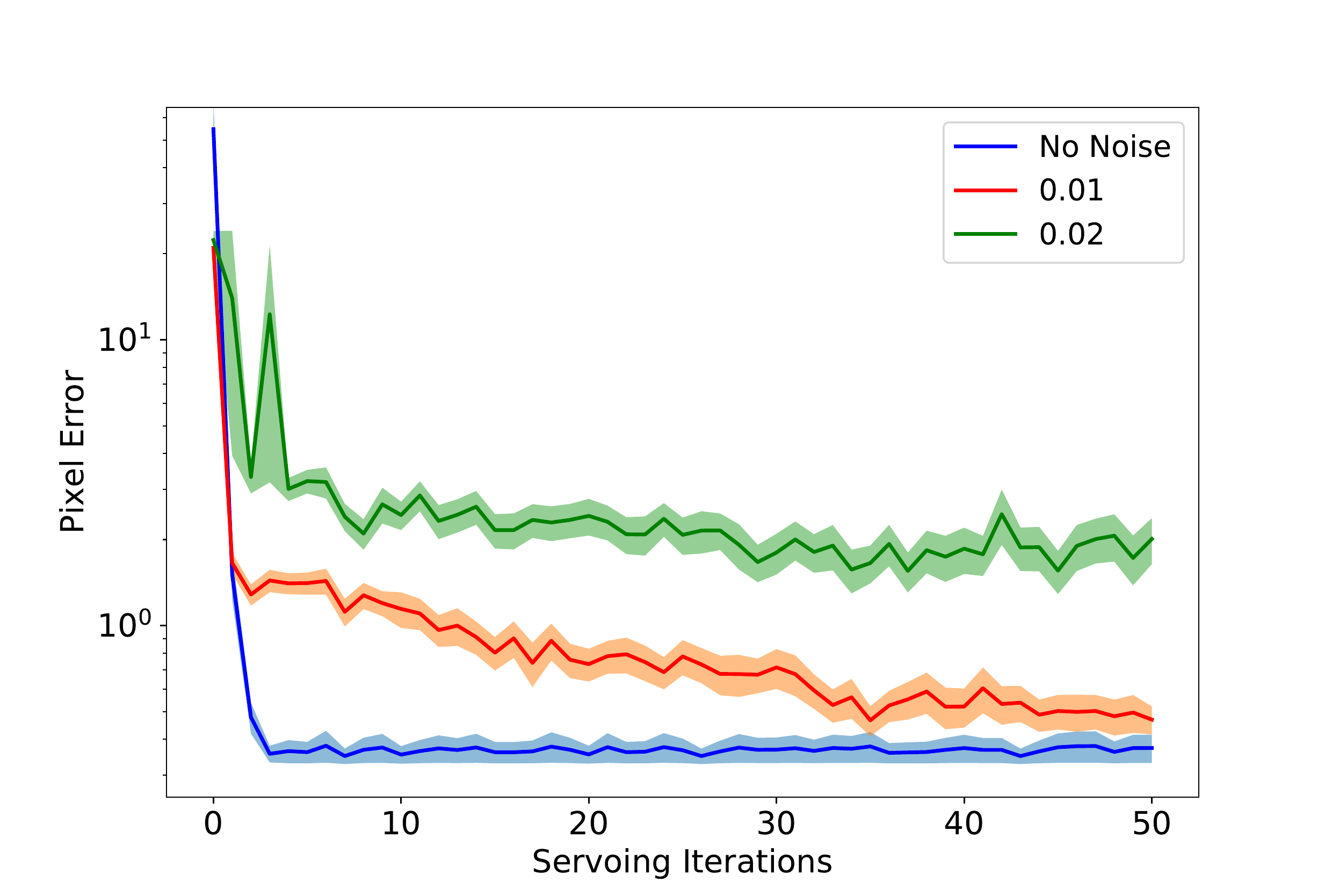}\caption{Pixel error vs. number of servoing iterations when operating a noisy Cartesian controller}\label{fig:cartnoise}\end{figure}

\begin{figure}\centering\includegraphics[width=\linewidth]{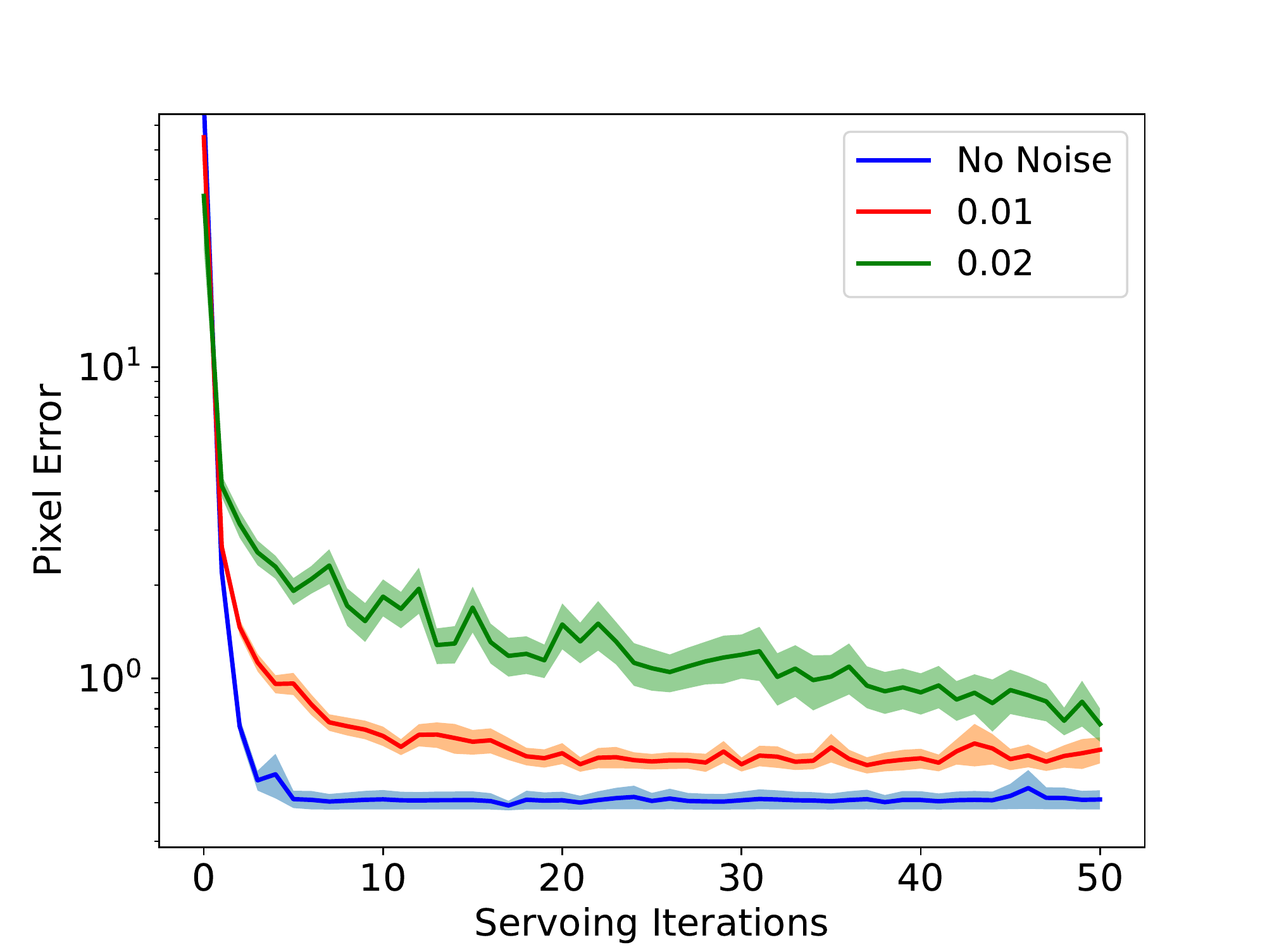}\caption{Pixel error vs. number of servoing iterations when operating a noisy joint controller}\label{fig:jointnoise}\end{figure}

In Figure \ref{fig:cartnoise}, we show the pixel error as a function of the number of servoing steps when operating a Cartesian controller with noise. Noise is simulated by adding Gaussian noise to the Euler coordinates of the action, which is a 6D transformation. Our model is trained by sending random actions to the noisy Cartesian controller and capturing images. We use 50 samples for learning. As the noise in the controller increases, the accuracy and speed of convergence decreases. However we are still able to position with pixel accuracy when the noise has a standard deviation $0.01$ meters in translation and $0.01$ radians in roll,pitch, and yaw. This becomes approximately $5$ pixel accuracy when the noise has a standard deviation of $0.02$ (meters and radians).

imgs/In Figure \ref{fig:jointnoise}, we show the pixel error as a function of the number of servoing steps when operating a joint controller with noise. Noise is simulated by adding Gaussian noise to the joint deltas sent to robot. Our model is  trained by sending random actions to the noisy joint controller and capturing images. We use 50 samples for learning. Even with the standard deviation of noise at $0.01$ radians and $0.02$ radians, we can still converge with pixel accuracy.

\section{Demonstration}
\label{sec:demo}

\begin{figure}[!ht]
\setlength{\fboxsep}{0pt}%
\setlength{\fboxrule}{0pt}%
\begin{center}
\includegraphics[width=0.9\linewidth]{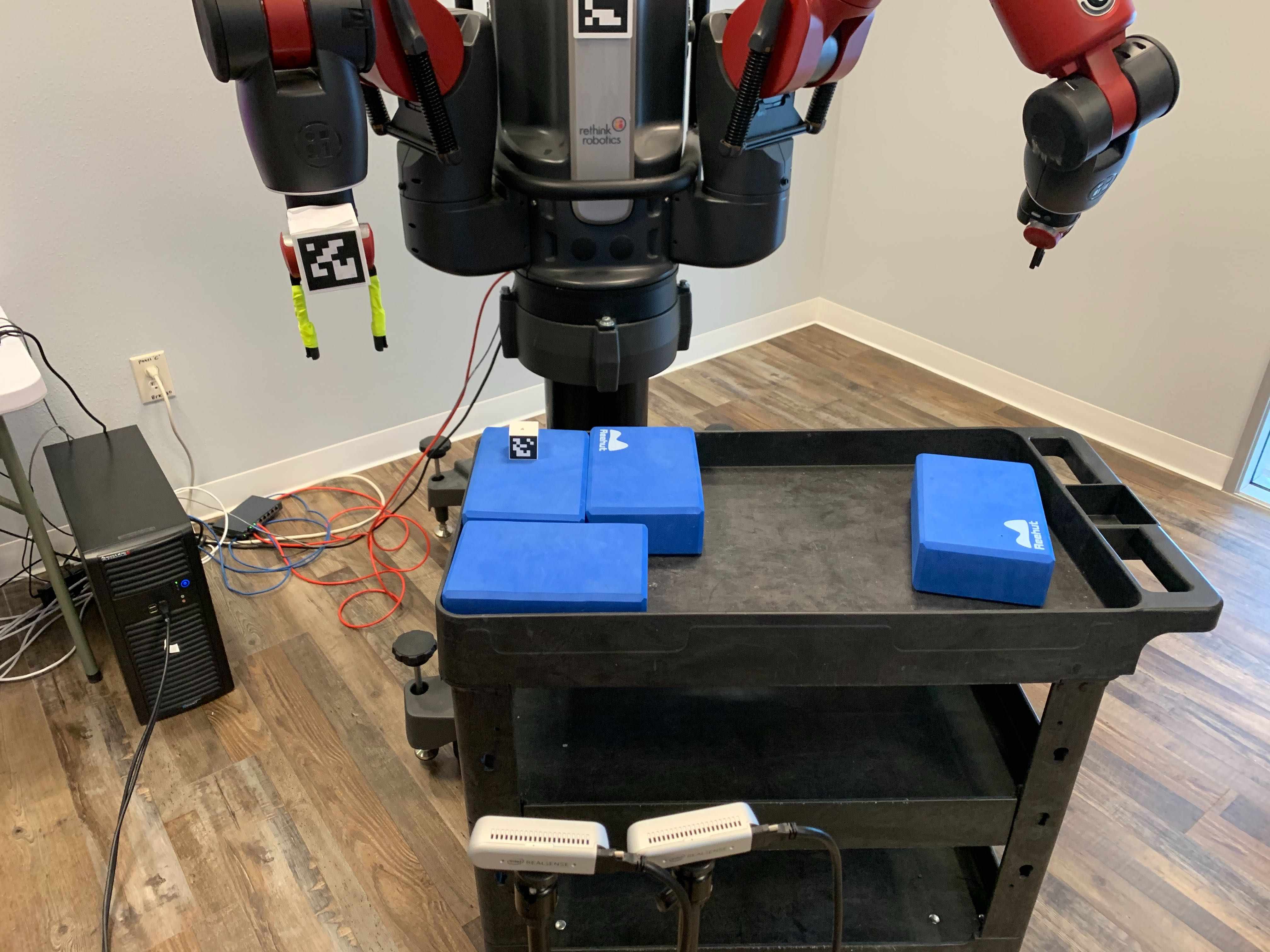}

\vspace{0.5cm}

\includegraphics[width=0.9\linewidth]{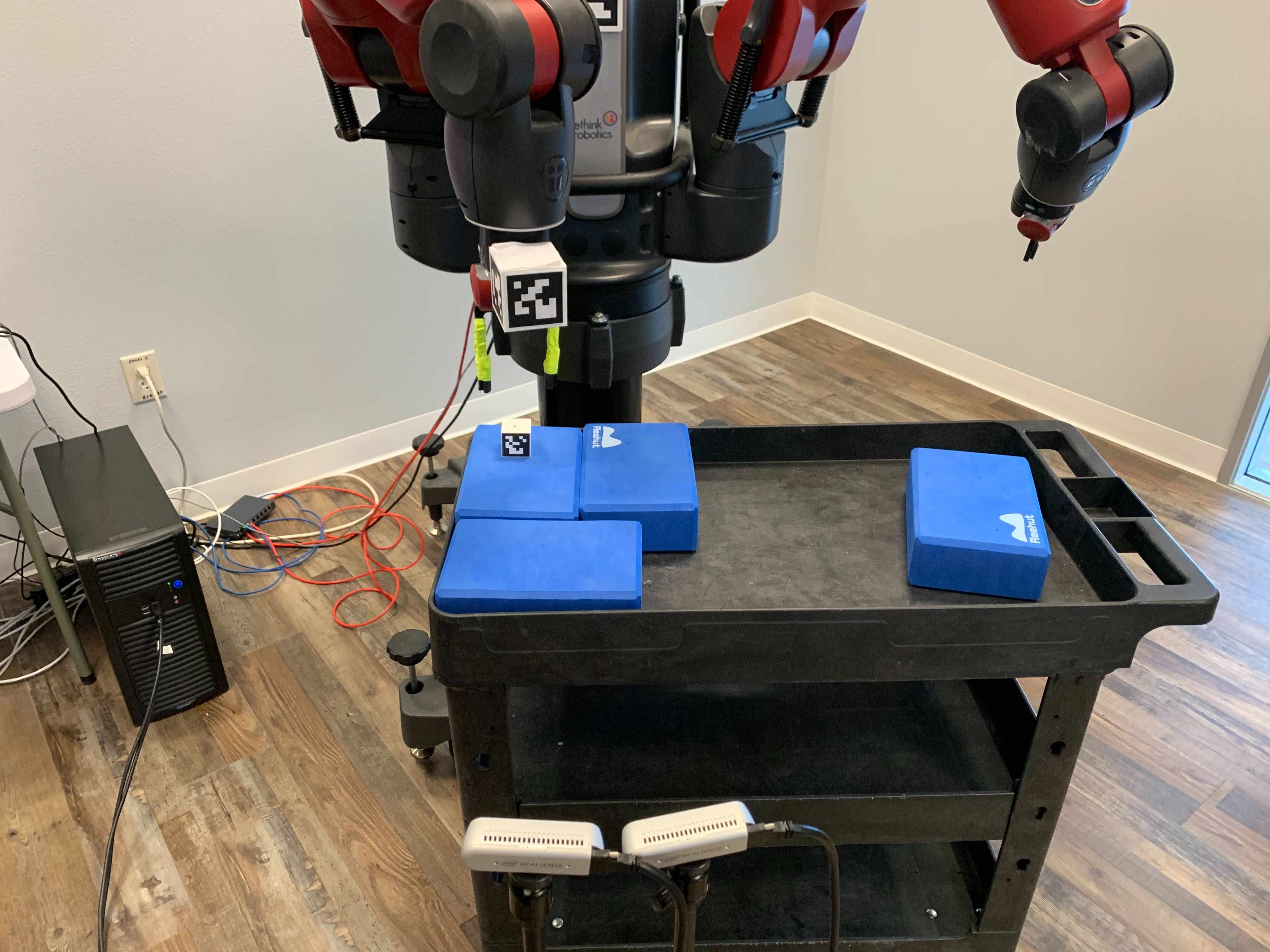}

\vspace{0.5cm}

\includegraphics[width=0.9\linewidth]{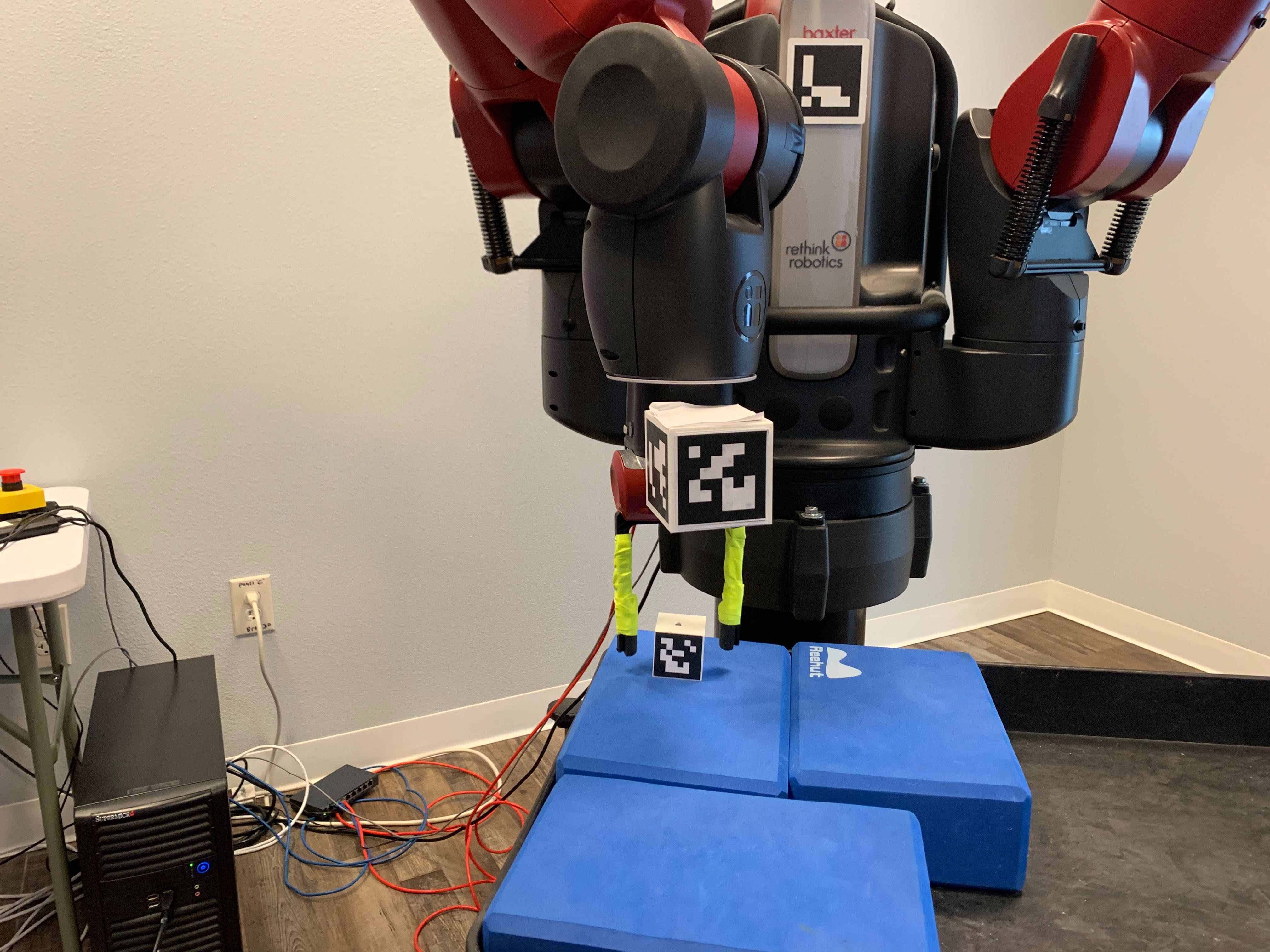}
\end{center}
\caption{Grasping}
\label{fig:grasping}
\end{figure}

\begin{figure}[!ht]
\setlength{\fboxsep}{0pt}%
\setlength{\fboxrule}{0pt}%
\begin{center}
\includegraphics[width=0.9\linewidth]{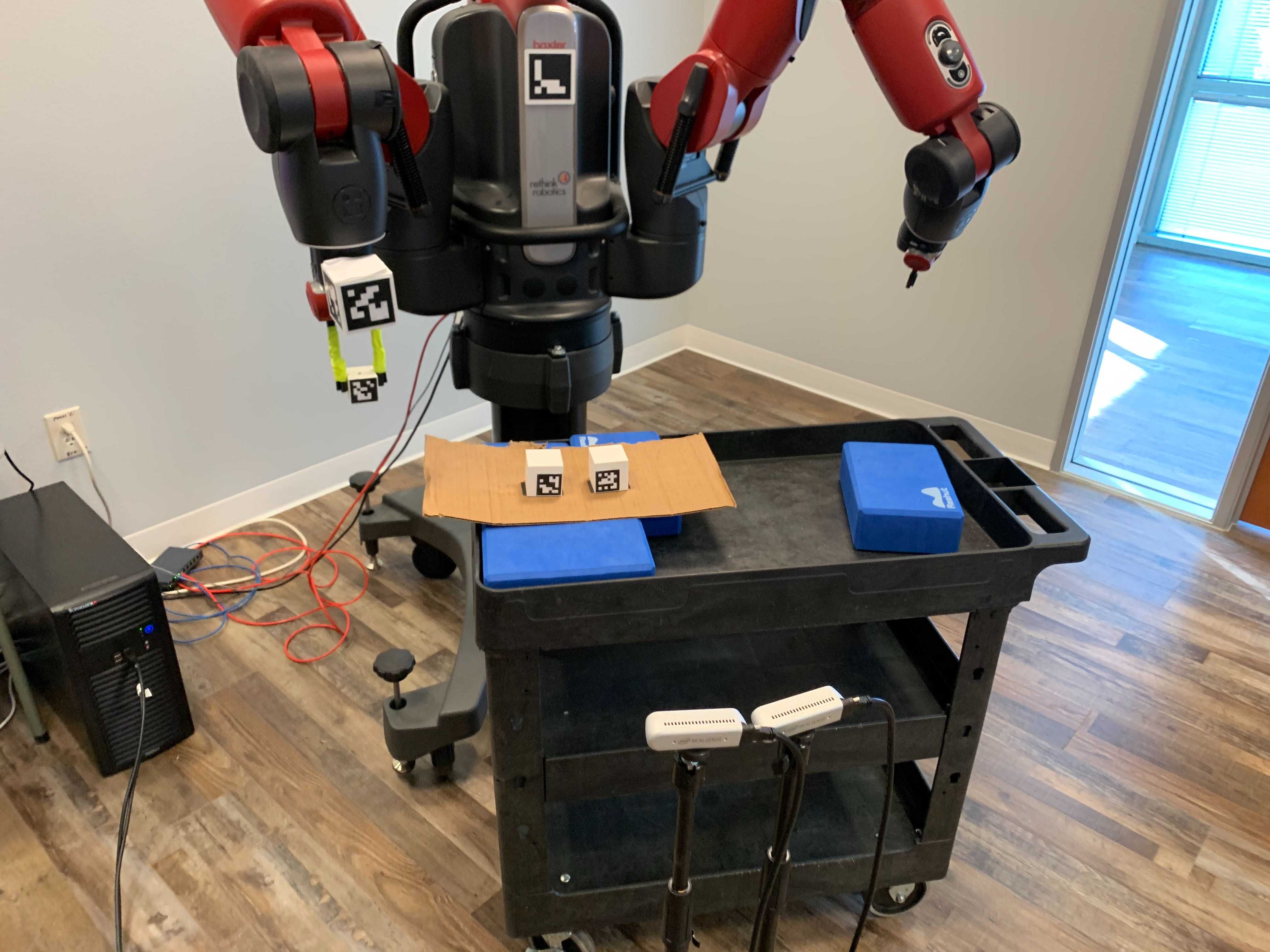}

\vspace{0.5cm}

\includegraphics[width=0.9\linewidth]{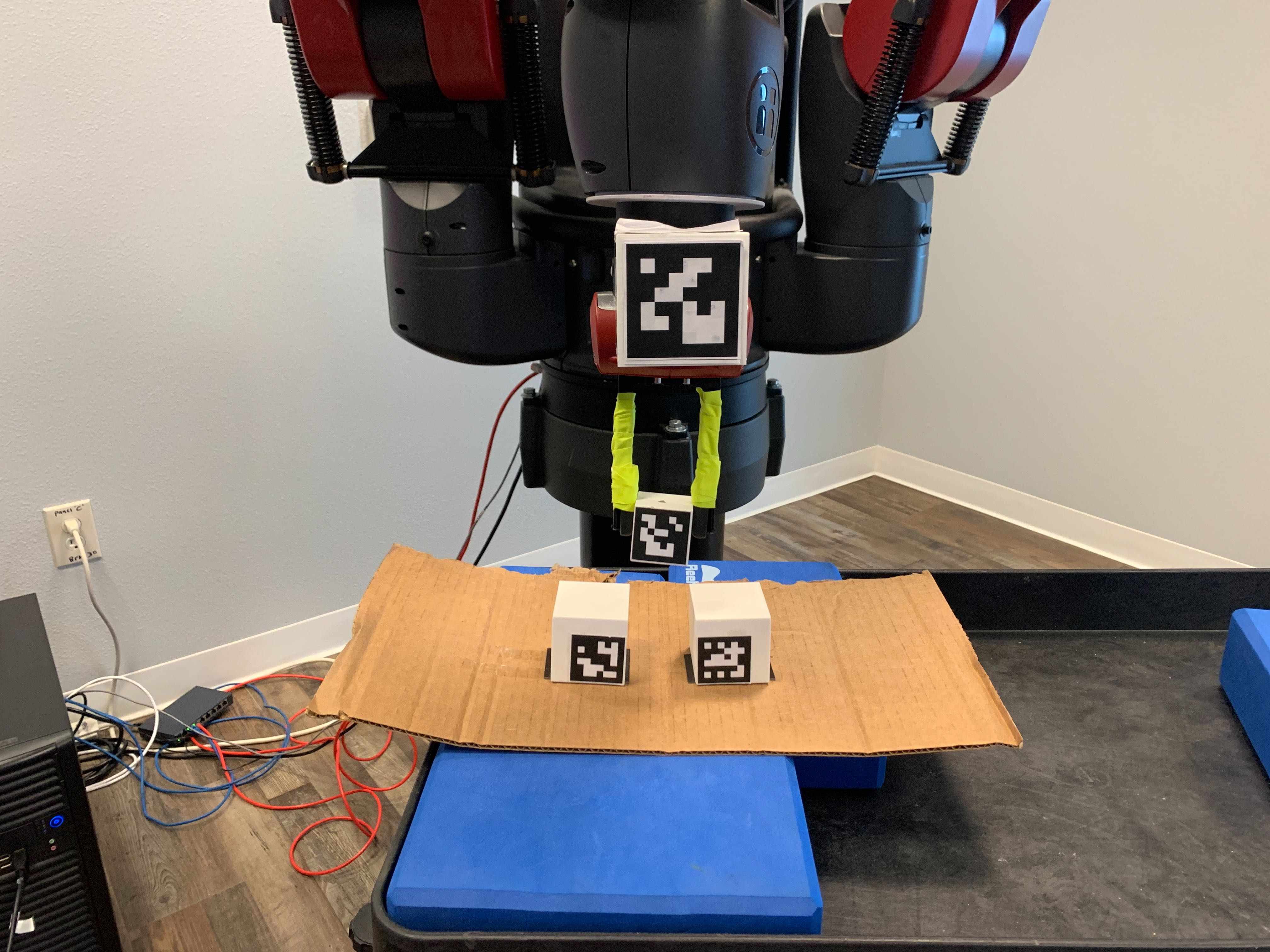}

\vspace{0.5cm}

\includegraphics[width=0.9\linewidth]{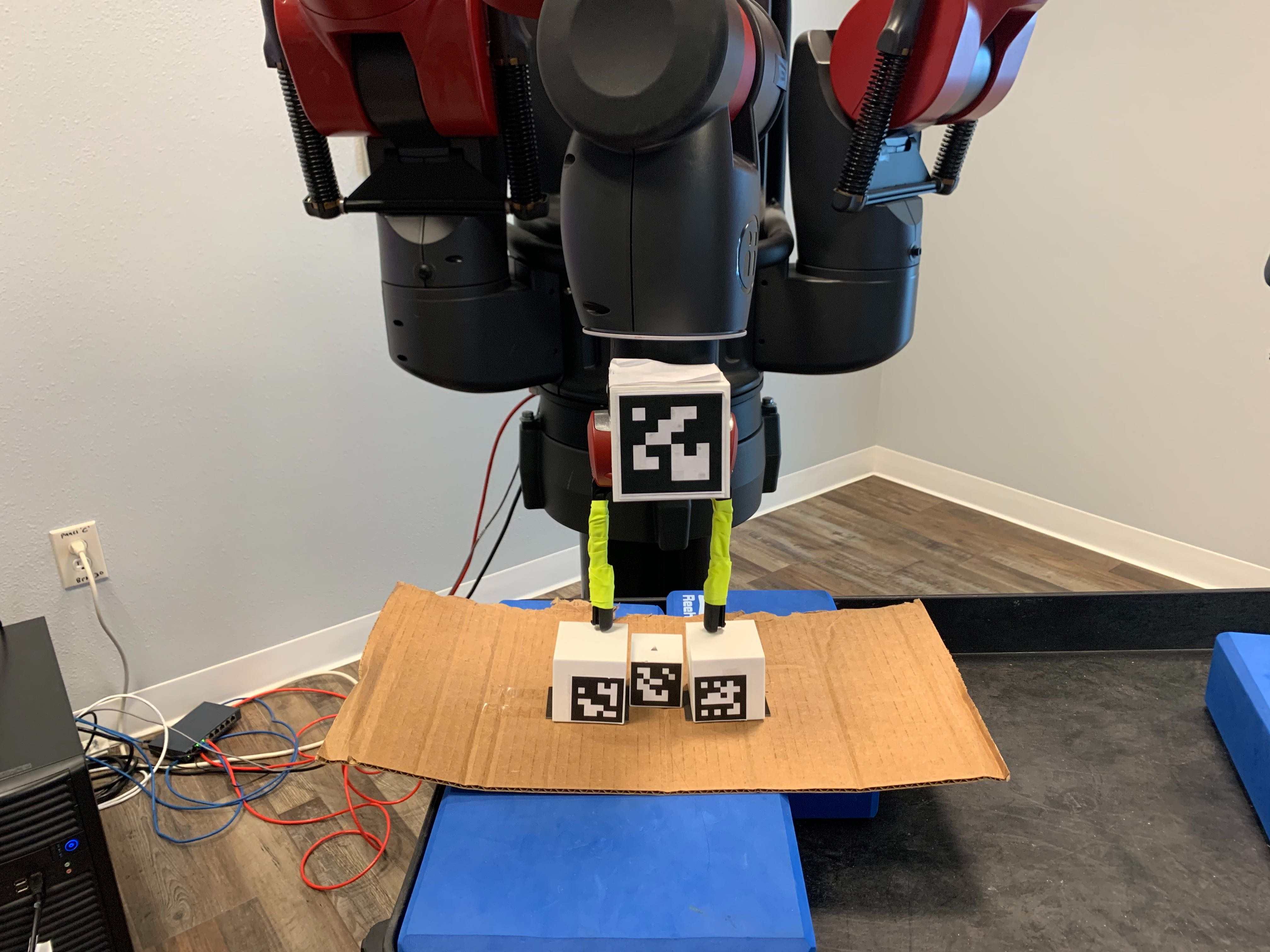}
\end{center}
\caption{Tight-fit Insertion}
\label{fig:insertion}
\end{figure}

We demonstrate our visual servoing method by executing grasping and tight-fit insertion motions on a Baxter robot. The Baxter robot uses series elastic actuators (SEAs) in their joints, which makes control inaccurate. We learn the model by moving to 50 random joint configurations.

These demonstrations can be produced using any of the learning methods we have described in this paper.

\subsection{Grasping}
In order to learn to grasp a new object, we must first learn its structure. To do this we place the object in the robot's gripper and collect 5 samples, which is sufficient to learn the object's structure if there is sufficient variety of views. Even if the structure is not entirely accurate, by running online learning we will eventually converge to the correct structure. After learning, we should be able to grasp that object located anywhere in view of a camera.

To perform the grasp, we compute the joint states corresponding to the object being inside the gripper. Since we learned the structure parameters with the object in the gripper, we can do this by considering only the features on the object and inferring the corresponding joint states.

Then given this target state we can iteratively converge to that target state. We do this by using only the features on the arm to infer the current joint states and then sending the delta between the current and target joint states.

However, moving directly to the grasping pose might cause the gripper to collide with the object. Instead we may want to approach the object from above. We do this by modifying the learned coordinates of the structure. We stretch them, increasing the distance between the object features and arm features. As we incrementally reduce this distance back to its original distance, doing inference of the joint states at each step, the gripper will move down to grip the object.

This procedure can allow for grasping with very few motions. For example we have found that grasping can be done consistently by taking 4 attempts to move to a pose computed after stretching the structure by a factor of 1.4. And then a single motion to stretch factors of 1.3, 1.2, 1.1, and finally 1.0.  Figure \ref{fig:grasping} shows the steps of executing a successful grasp.

\subsection{Insertion}

Now once we have successfully grasped the object, we execute a tight-fit insertion. A high-precision insertion will need to be done visually since small errors in the pose estimation can cause the insertion to fail.

First we need to learn the structure of the insertion target. As shown in Figure \ref{fig:insertion}, we mark the insertion target using two markers that signify the edges of the target location. Once we learn this, we can place that target anywhere in view of the image (or move the cameras and leave the target fixed) and we will be able to properly identify the target location.

Additionally, since the object may have been grasped differently than how we originally learned, we can relearn the structure of the in-hand object. Again this can be done with only 5 samples.

Then to do the insertion, we infer the target joint states corresponding to the target pixel locations of the object features (at the insertion target). Then we infer our current state by using the arm features.  We command the delta between the current and target. As is done for grasping, to gradually approach the insertion target we can gradually scale the structure features. Figure \ref{fig:insertion} shows the steps of executing a successful insertion.

\section{Conclusion}
\label{sec:conclusion}

In this work, we presented a generative, model-based approach of learning to control a robot using visual feedback. Our insight is that we can avoid calibration of the true, physical kinematic and camera parameters and instead just directly learn a model to generate feature pixel coordinates from joint states and actions. With our training procedure, we show that it is feasible to do so with no prior information about the system. Training the model is done sample-efficiently, with as little as 50 observations, and shows strong generalization to test data sets. The model explicitly handles uncertainty of the observations, which allows us to control robots that provide noisy measurements of the joint states or Cartesian pose. Further, the learning is feasible even when the robot state is not observed, only the actions that we command to a noisy controller. Our modular formulation enables us to flexibly add and relearn components. We showed that we can quickly learn to use new cameras and understand the structure of new objects to manipulate them. Through online learning, our model is robust to changes in the system since it can quickly adapt its parameters. Inference in the model enables us to infer the robot's state given target coordinates for the features in the image. We demonstrated that our method allows matching target coordinates of the features in significantly fewer iterations than state-of-the-art methods in visual servoing. We concluded with demonstrations of grasping and insertion behaviors on a robot with a noisy controller.

We plan to release an implementation of our method to enable researchers to use and extend our system and build useful applications with it. Our method can be used in place of or as a sub-module within neural network architectures that attempt to learn control of a robot, which will provide them benefits in terms of sample-efficiency, speed, and accuracy. We are excited about the future research directions, particularly in manipulation and control, that our work lays the foundation for.

\bibliographystyle{SageH}
\bibliography{main.bib}

\end{document}